\documentclass{article}

\usepackage{arxiv}

\usepackage[utf8]{inputenc} 
\usepackage[T1]{fontenc}    
\usepackage{hyperref}       
\usepackage{url}            
\usepackage{booktabs}       
\usepackage{amsfonts}       
\usepackage{nicefrac}       
\usepackage{microtype}      
\usepackage{lipsum}		
\usepackage{graphicx}
\usepackage{natbib}
\usepackage{doi}
\usepackage{amssymb}
\usepackage{longtable, booktabs}
\usepackage{xcolor}
\usepackage[color=green]{todonotes}  
\usepackage{blindtext}
\usepackage{algorithm, algorithmicx, algpseudocode}
\usepackage{amsmath}
\usepackage{multirow}
\usepackage{graphicx}
\usepackage{array}

\title{A Fast Fourier Convolutional Deep Neural Network For Accurate and Explainable Discrimination Of Wheat Yellow Rust And Nitrogen Deficiency From Sentinel-2 Time-Series Data}

\author{ \href{https://orcid.org/0000-0001-8424-6996}{\includegraphics[scale=0.06]{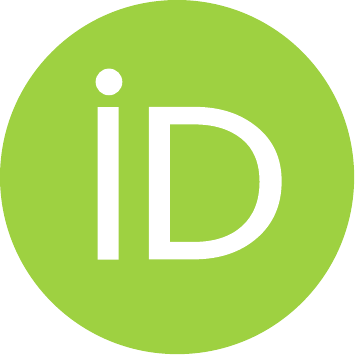}\hspace{1mm}Yue~Shi}\\
	School of Electronic and Electrical Engineering, \\
	The University of Leeds, \\
	Leeds, UK \\
	\texttt{y.shi1@leeds.ac.uk} \\
	\And
	\href{https://orcid.org/0000-0000-0000-0000}{\includegraphics[scale=0.06]{orcid.pdf}\hspace{1mm}Liangxiu~Han} \\
	Department of Computing and Mathematics\\
	Manchester Metropolitan University\\
	Manchester M1 5GD, UK \\
	\texttt{L.han@mmu.ac.uk} \\
	\And
	\href{https://orcid.org/0000-0000-0000-0000}{\includegraphics[scale=0.06]{orcid.pdf}\hspace{1mm}Pablo~González-Moreno} \\
	Department of Forest Engineering\\
	University of Córdoba\\
	Córdoba, Spain\\
    \And
	\href{https://orcid.org/0000-0000-0000-0000}{\includegraphics[scale=0.06]{orcid.pdf}\hspace{1mm}Darren~Dancy} \\
	Department of Computing and Mathematics\\
	Manchester Metropolitan University\\
	Manchester M1 5GD, UK \\
    \And
	\href{https://orcid.org/0000-0000-0000-0000}{\includegraphics[scale=0.06]{orcid.pdf}\hspace{1mm}Wenjiang~Huang} \\
	Aerospace Information Research Institute\\
	Chinese Academy of Sciences, \\
    Beijing 100094, China.\\
    \And
	\href{https://orcid.org/0000-0000-0000-0000}{\includegraphics[scale=0.06]{orcid.pdf}\hspace{1mm}Zhiqiang~Zhang} \\
	School of Electronic and Electrical Engineering, \\
	The University of Leeds, \\
	Leeds, UK \\
    \And
	\href{https://orcid.org/0000-0000-0000-0000}{\includegraphics[scale=0.06]{orcid.pdf}\hspace{1mm}Yuanyuan~Liu} \\
	Department of Computer Science, \\
	The University of Manchester, \\
	Manchester, UK \\
    \And
	\href{https://orcid.org/0000-0000-0000-0000}{\includegraphics[scale=0.06]{orcid.pdf}\hspace{1mm}Mengning Huang} \\
	Beijing University of Technology \\
	Beijing, China \\
    \And
	\href{https://orcid.org/0000-0000-0000-0000}{\includegraphics[scale=0.06]{orcid.pdf}\hspace{1mm}Hong Miao} \\
	College of Mechanical Engineering,  \\
	Yangzhou University, \\
    Yangzhou,  China \\
    \And
	\href{https://orcid.org/0000-0000-0000-0000}{\includegraphics[scale=0.06]{orcid.pdf}\hspace{1mm}Min Dai} \\
	College of Mechanical Engineering,  \\
	Yangzhou University, \\
    Yangzhou,  China \\ 
}

\renewcommand{\shorttitle}{\textit{arXiv} Template}

\hypersetup{
pdftitle={A template for the arxiv style},
pdfsubject={q-bio.NC, q-bio.QM},
pdfauthor={David S.~Hippocampus, Elias D.~Striatum},
pdfkeywords={First keyword, Second keyword, More},
}

\begin{document}
\maketitle

\begin{abstract}
	Accurate and timely detection of plant stress is essential for yield protection, allowing better-targeted intervention strategies. Recent advances in remote sensing and deep learning have shown great potential for rapid non-invasive detection of plant stress in a fully automated and reproducible manner.  However, the existing models always face several challenges: 1) computational inefficiency and the misclassifications between the different stresses with similar symptoms; and 2) the poor interpretability of the host-stress interaction.  In this work, we propose a novel fast Fourier Convolutional Neural Network (FFDNN) for accurate and explainable detection of two plant stresses with similar symptoms (i.e. Wheat Yellow Rust And Nitrogen Deficiency). Specifically, unlike the existing CNN models, the main components of the proposed model include: 1) a fast Fourier convolutional block, a newly fast Fourier transformation kernel as the basic perception unit, to substitute the traditional convolutional kernel to capture both local and global responses to plant stress in various time-scale and improve computing efficiency with reduced learning parameters in Fourier domain; 2) Capsule Feature Encoder to encapsulate the extracted features into a series of vector features to represent part-to-whole relationship with the hierarchical structure of the host-stress interactions of the specific stress. In addition, in order to alleviate over-fitting, a photochemical vegetation indices-based filter is placed as pre-processing operator to remove the non-photochemical noises from the input Sentinel-2 time series. The proposed model has been evaluated with ground truth data under both controlled and natural conditions. The results demonstrate that the high-level vector features interpret the influence of the host-stress interaction/response and the proposed model achieves competitive advantages in the detection and discrimination of yellow rust and nitrogen deficiency on Sentinel-2 time series in terms of classification accuracy, robustness, and generalization.
\end{abstract}


\section{Introduction}

\label{sec:1}

The plant stress caused by unfavorable environmental conditions (e.g. a lack of nutrients, insufficient water, disease, or insect damage), if left untreated, will lead to irreversible damage and decreases in plant production. Early accurate detection of plant stress is essential to be able to respond with appropriate interventions to reverse stress and minimize yield loss. 
Recent advances in remote sensing with enhanced spatial, temporal and spectral capacities, combined with deep learning have offered unprecedented possibilities for rapid, non-invasive stress detection in a fully automated and reproducible manner \citep{ji20183d,wang2020deep}.  
Currently, the deep learning models have been proven effective in remote sensing time-series analysis of plant stresses \citep{RN73, RN71}. 1D-CNN and 2D-CNN with convolutions were applied either in the spectral domain or in the spatial domain \citep{scarpa2018cnn, kussul2017deep}. In addition, 3D-CNNs were also used across spectral and spatial dimensions \citep{li2017spectral, hamida20183}. These models do not consider temporal information.  Meanwhile, temporal 1D-CNNs were proposed to handle the temporal dimension for general time series classification \citep{wang2017time} and RNN-based models to extract features from multi-temporal observation by leveraging the sequential properties of multi-spectral data and combination of RNN \citep{kamilaris2018deep} and 2D-CNNs where convolutions were applied in both temporal and spatial dimensions \citep{zhong2019deep}. These preliminary works highlight the importance of temporal information which can improve the classification accuracy performance.  Despite the existing works are encouraging, it suffers several limitations: 1) Over-fitting and uncertainty caused by noisy data involved in the remote sensing time-series; 2) Computing inefficiency and inaccuracy caused by the convolutional operations that are applied to all layer, particularly with the increase of size of images and the kernel. In particular, for multi-plant stresses classification, similar symptoms always lead to confusion during classification as most of the local features are extracted from the neighbor time steps. Therefore, a more effective denoise operator and larger receptive fields for the extraction of the global biological responses at various time-scale are highly desired.  \par
 
One solution is to pre-filter the photochemical information from satellite time-series and change the domain through Fourier transform to model the part-to-whole relationship between the photochemical features and specific plant stress in the frequency domain. This is because the convolution operation in the spatial domain is the same as the point-by-point multiplication in the Fourier domain.  According to the Fourier theory, Fourier transform provides an effective perception operation with non-local receptive field. Unlike existing CNNs where a large-sized kernel is used to extract local features, Fourier transforms with a small-sized kernel can capture global information. Therefore, the Fourier kernel has a great potential in replacing the traditional convolutional kernel in remote sensing time-series analysis without any additional effort \citep{yi2023neural}. For example, \cite{chen2023fourier} designed a Fourier domain structural relationship analysis framework to exploits both modality-independent local and nonlocal structural relationships for unsupervised change detection. However, the existing Fourier operators can only be sparsely inserted into the deep learning network pipeline due to their expensive computational cost. Therefore, the Fast Fourier Transformation (FFT) is an effective way to extract the global feature responses from satellite image time-series \citep{RN76}. For example, \cite{awujoola2022multi} proposed a multi-stream fast Fourier convolutional neural network (MS-FFCNN) by utilizing the fast Fourier transformation instead of the traditional convolution, it lowers the computing cost of image convolution in CNNs, which lowers the overall computational cost. \cite{lingyun2022spectral} designed a spectral deep network combining Fourier Convolution (FFC) and classifier by extending the receptive field. Their results demonstrated that the features around the object provide the explainable information for small objects detection.  \par

Although the effectiveness of Fourier-based convolution has been proved by many studies, few studies have done in the multiple plant stress detection from remote sensing data. In this work, we have proposed a novel fast Fourier convolutional deep neural network (FFCDNN) for accurate and early efficient detection of plant stress with initial focus on wheat yellow rust (\textit{Puccinia striiformis}) and nitrogen deficiency. The proposed model significantly reduce the computing cost with improved accuracy and interpretability.  Specifically, a newly fast Fourier transformation (FFT) kernel is proposed as the basic perception unit of the network to extract the stress-associated biological dynamics with various time-scale; and then the extracted biological dynamics are encapsulated into a series of high-level featured vectors representing the host-stress interactions specific to different stresses; finally, a non-linear activation function is designed to achieve the final decision of the classification. The proposed model has been evaluated with ground truth data under both the controlled and natural conditions. \par

The rest of this work is organized as follows. Section 2 introduces related works on existing methods of multiple plant disease classification; Section 3 details the proposed approach. Section 4 presents the material and experiment details. Section 5 illustrates the experimental evaluation results. Finally, Section 6 concludes the work.

\section{The related work}

\subsection{Plant photochemical information filter from satellite images}
The newly launched satellite sensors (e.g. Sentinel-2, worldview-3, etc.) provide the promising EO dataset for improved plant photochemical estimation \citep{RN14}. Wherein, leaf chlorophyll content (LCC), canopy chlorophyll content (CCC), and leaf area index (LAI) are the most popular remotely retrievable indicators for detecting and discriminating plant stresses \citep{RN45, RN43}. Among these indicators, the LCC time-series is a key biochemical dynamics for the stresses-associated foliar component changes without (or partly) the effects from soil background and canopy structure. Estimating LCC requires the remote sensing indicators that are sensitive to the LCC but, at the same time, is insensitive to LAI and background effects \citep{RN45}. On the other hand, the LAI is one of the critical biophysics-specific proxies in characterizing the canopy architecture variations that response to the apparent symptom caused by specific stress \citep{RN32}. By contrast, CCC is determined by the LAI and LCC, expressed per unit leaf area, which remains multicollinearity with them and hard to be used in separating the stresses-induced biochemical changes from the biophysical impacts. Therefore, the LCC and LAI are regarded as a pair of independent variables for filtering the biochemical information between the different plant stresses \citep{RN12, RN10}. \par

Regarding to the filter methods, by using the reflectance in red-edge regions, there are two methods used in LAI and LCC estimation for minimizing saturation effect and soil background associated noises: 1) the vegetation index method \citep{RN18, RN32} and 2) the radiative transfer models (RTM) \citep{RN77, RN78}. For example, \cite{RN22} tested and compared the performance of the red-edge chlorophyll index (CIred-edge) and green chlorophyll index (CIgreen) on the Sentinel-2 bands, their results indicated that the setting of Sentinel-2 bands is well positioned for deriving these indices on LCC estimation. \cite{RN74} combined developed a PROSAIL-based model to estimate LAI and biomass on the Sentinel-2 bands, and the yielded LAI values are in agreement with the ground truth LAI measurements. However, the simply use of the remotely estimated LAI and LCC is hard to represent the non-linear host-stress interactions of the plant stresses.  \par

\subsection{Plant stress detection methods}
Currently, there are two types of methods widely used in extracting the interpretable agent features for the plant stresses from satellite imagery, including the biological methods and the deep learning-based methods. \par

\subsubsection{Biological methods}
Studies have shown that biological models can be used to map within-field crop stresses variability \citep{zhou2021diagnosis, ryu2020performances}. This is possible because the infestation of crop stresses often leads plants to close their stomata, decreasing canopy stomatal conductance and transpiration, which in turn raises foliar biophysical and biochemical variations \citep{tan2019sensitivity}. 
However, plant stress involves the complicated biophysical and biochemical responses, which demands the unique biological index only partially represent the crop stress. For instance, leaf area index (LAI) is a direct indicator of plant canopy structure features \citep{ihuoma2019sensitivity}. The stressed plants will lead to fluctuations on plants LAI time-series with different patterns, which will raise the higher radiations of a stressed crop \citep{ballester2019monitoring}. \cite{jiang2020simulating} proposed two LAI-derived soil water stress functions in order to quantify the effect of soil water stress on the processes of leaf expansion and leaf senescence caused by the stresses. Their results showed that the LAI-based model is sensitive to the stress-derived leaf expansion. \cite{zhu2021relating} developed a vegetation indices derived model from the observed hyperspectral data of winter wheat to detect the plant salinity, the results show that the salt-sensitive blue, red-edge, and near-infrared wavebands have great performances on the detection of the plant salinity stress.\par

Unlike the LAI, the Photochemical associated indices directly account for leaf physiological changes such as photosynthetic pigment changes \citep{gerhards2019challenges}. Photochemical reflectance is the dominant factor determining leaf reflectance in the visible wavelength ($400 - 700$ nm), with chlorophyll considered the most relevant photochemical index for crop stress diagnosis \citep{zhou2021assessment}. Under prolonged infestations, leaf chlorophyll content often decreases, leading to a reduction in green reflection and an increase in blue and red reflections. The spectral radiation characteristics between the red and near-infrared regions are sensitive to leaf and canopy chlorophyll content. The ratio of red and near-infrared has shown a strong sensitivity to the crop stress associated chlorophyll content changes \citep{ryu2020performances}. \cite{cao2019comparison} compared the feasibility of the leaf chlorophyll content, net photosynthesis rate, and maximum efficiency of photosystem on the detection crop heat stress, their findings suggest that the maximum efficiency of photosystem was the most sensitive remote sensing agent to heat stress and had the ability to indicate the start and end of the stress at the slight level or the early stage. \cite{shivers2019using} used the visible-shortwave infrared (VSWIR) spectra to model the non-photosynthetic vegetation and soil background from the Airborne Visible/Infrared Imaging Spectrometer (AVIRIS), their finding revealed the increase in temperature residuals is highly consistent with the infestation of crop stresses.\par

\subsubsection{Machine/deep learning-based methods}
Despite many studies have been focusing on crop stress detection using the biological characteristics, most of the applications require self-adjusted algorithms to improve the robustness and generalization of the model for the complicated nature conditions. Among the crop stress detection techniques, machine learning and deep learning have played a key role. For machine learning approaches, supervised models have been proved effective in data mining from the training dataset \citep{kaneda2017multi}. The data flow in the machine learning models includes feature extraction, data assimilation, optimal decision boundary searching, and classifiers for stress diagnosis. Wherein, supervised learning is deal with the classification issues by representing the labeled samples. Such models aim to find the optimal model parameters to predict the unlabelled samples \citep{harrington2012machine}.\par

Deep learning has many neural layers which transforms the sensitive information from the input to output (i.e., healthy or stressed). The most applied perception neural unit is convolutional neural unit in crop stress detection \citep{fuentes2017robust, krishnaswamy2020disease}. Generally, the convolutional neural unit consists of dozens of layers that process the input information with convolution kernel. In the area of crop stress detection, deep learning contributed significantly to the analysis of plant stress high-level features \citep{jin2018classifying}. In crop stress image classification, the multi-source images are usually used as input to extract the stress dynamics during their developments, and a diagnostic decision is used as output (e.g., healthy or diseased) \citep{an2019identification, cruz2019detection}. \cite{barbedo2019plant} developed a convolutional deep learning model to classify individual lesions and spots on plant leaves. This model has been successfully used in the identification of multiple diseases, the accuracy obtained in this model was $12\%$ higher than that traditional models. \cite{lin2019deep} applied a convolutional kernel based U-Net to segment cucumber powdery mildew-infected cucumber leaves, the proposed binary cross entropy loss function is used to magnify the loss of the powdery mildew stressed pixels, the average accuracy for the powdery mildew detection reaches $96.08\%$.

\subsection{Interpretability of deep learning-based models}
Although the deep learning models have been successfully applied for vegetation stress monitoring applications, most of existing deep learning-based approaches have difficulty in explaining plant biophysical and biochemical characteristics due to their black-box representations of the features extracted from intermediate layers \citep{shi2020biologically}. Thus, the interpretability of deep models has become one of the most active research topics in the remote sensing based crop stress diagnosis, which can enhance and improve the robustness and accuracy of models in the vegetation monitoring applications from the biological perspective of target entities \citep{too2019comparative, brahimi2019deep}. \par

Recently, the model interpretability used to disclose the intrinsic learning logic for detection and discrimination of plant stresses has received growing attention \citep{RN2015}. In other words, the interpretability that illustrates the performance of the model on characterizing the specific host-stress interaction guarantees the generalization ability of the model for the practise usages. Among the existing models, the visulization of the feature representation is the most direct method for improving the model interpretability. For example, \cite{behmann2014detection} proposed an unsupervised model for early detection of the drought stress in barley, wherein, the intermediate features produced by this model highly related with the sensitive spectral bands for drought stress. Another way to improve the interpretability of deep learning models is to construct the network architecture which can bring the network an explicit semantic meaning. For example, \cite{shi2020biologically} developed a biologically interpretable two-stage deep neural network (BIT-DNN) for detection and classification of yellow rust from the hyperspectral imagery. Their findings demonstrate that the BIT-DNN has great advantages in terms of the accuracy and interpretability.  \par

\subsection{Fast Fourier Transformation} 

The traditional receptive field used in convolution operations only conduct to the central regions to extract the local features related to the interested targets. This limits the necessity of large convolutional kernel on global feature extraction. Recently, 
there is an increasing interest in applying Fourier transform to deep nural networks to capture global feature. as mentioned in the introduction section, Fourier transform provides an effective perception operation with non-local receptive field. Unlike existing CNNs where a large-sized kernel is used to extract local features, Fourier transform with a small sized kernel is able to capture global information. For example, \citep{rippel2015spectral} proposed Fourier transformation pooling layer that performs like principle component extraction by constructing the representation in the frequency domain. \citep{chi2019fast} proposed a integrated the Fourier transforms into a series of convolution layers in the frequency domain. \par

Fast Fourier Transformation-based deep learning models use the time-frequency analysis methods to extract the low-frequency host-stress interaction by limiting the high-frequency noises in the frequency-domain space \citep{RN30, RN31, RN46, RN40}. FFT is an useful harmonic analysis tool, which has been widely used in reconstruction of vegetation index time-series \citep{RN49}, curve smoothing \citep{RN52, RN50}, and ecological and phenological applications \citep{RN53, RN54, RN51}. FFT maps the satellite time-series signals into superimposed sequences of cosines waves (terms) with variant frequencies, each component term accounting for a percentage of the total variance in the original time-series data \citep{RN46}. This process facilitates the recognition of subtle patterns of interest from the complex background noises, which degrade the spectral information required to capture vegetation properties\citep{RN67, RN68}. For example, \cite{RN66} used the fast Fourier transform (FFT) method to characterize temporal patterns of the fungal disease on winter wheat between the observation sites, and then achieved the fungal disease monitoring and forecasting at the regional level. Our work advances above-mentioned research front via designing an novel fast Fourier convolutional operation unit that simultaneously uses spatial and temporal information for achieving global feature extraction during the learning process. \par


\section{The proposed Fast Fourier Convolution Deep Neural Network (FFCDNN)}
To address the challenge of the misclassification for the different plant stresses with similar symptoms, we propose a novel Fast Fourier Convolution operator to efficiently implement non-local receptive field and fuses the extracted biological information with various time-scales in frequency domain, and then, a new deep learning architecture is developed to retrieve the host-stress interaction and achieve the high accurate classification. In this section, we describe the main framework of the proposed Fast Fourier Convolution Deep Neural Network (FFCDNN), in the context of multiple plant stresses discrimination from the agent-based biological dynamics.

\subsection{The network architecture of the proposed FFCDNN}
Fig.\ref{fig:5} depicts the main framework of the proposed FFCDNN for multiple crop stresses discrimination, in the context of Sentinel-2 derived biological agents (i.e. $VI_{LAI}$ and $VI_{LCC}$). To be specific, a branch structure is designed to respectively pre-filter the biochemical dynamics represented by $VI_{LAI}$ and $VI_{LCC}$ time-series. For each of the branch, the Fourier kernel is set as the same size as the input size of $VI_{LAI}$ and $VI_{LCC}$ time-domain (time-series) patches, with a size of $k \times k \times K^{(1)}$, then, the Fourier kernel is point-wised multiplied by the input biological agent patches. After the Fourier convolution is performed, the ReLU function is implemented to calculate the $VI_{LAI}$ and $VI_{LCC}$ time-series magnitude in frequency-domain containing stress-associated biological responses, and the activation feature map, with a size of $k \times k \times K^{(2)}$, is conducted with Fourier Pool layer to highlight the most important stress information and down sampling the feature map.

Subsequently, the $VI_{LAI}$ and $VI_{LCC}$ feature maps are sent to the hierarchical structure of the class-capsule blocks in order to build the part-to-whole relationship and to generate the hierarchical vector features for representing the high-level stress-pathogen interaction. Finally, a decoder is employed to predict the classes based on the length and direction of the hierarchical vector features in the feature space. The detailed information for the model blocks is described below:

\begin{figure}[]   
    \centering  
    \includegraphics[width=5.5in]{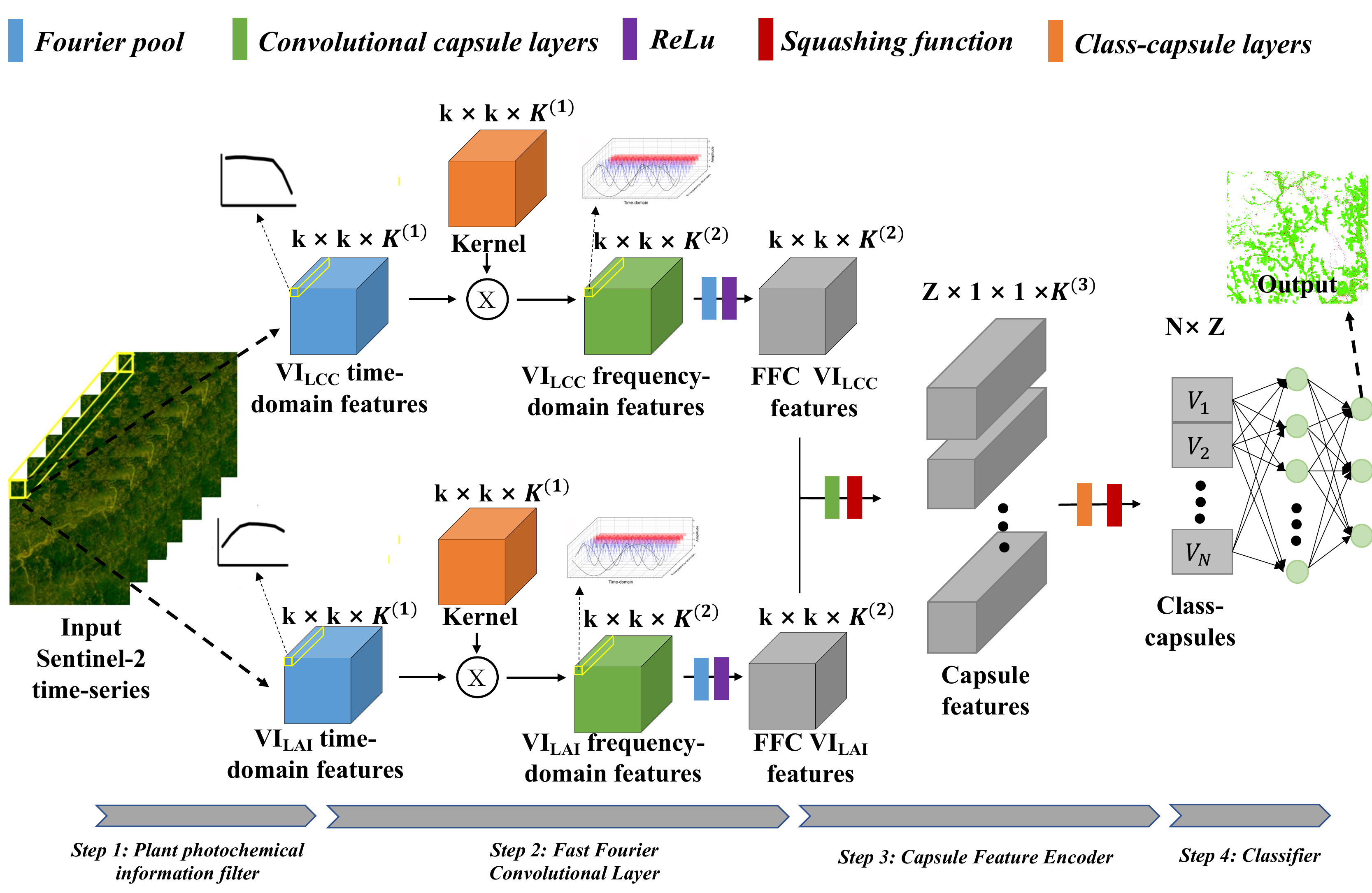}   
    \caption{ The workflow of the FFCDNN framework for the discrimination of multiple plant stresses from Sentinel-2 time-series}  
    \label{fig:5}  
\end{figure}

\textit{a.Plant photochemical information filter} \par
In this study, an agent-based photochemical information pre-filter is set as the pre-processing operator for the input satellite time-series. Based on the benchmark study of the existing vegetation agent models for LAI and LCC estimation shown in Appendix A, we use the Weighted Difference Vegetation Index (WDVI)-derived LAI, defined as $VI_{LAI}$, and TCARI/OSAVI-derived LCC, defined as $VI_{LCC}$, as the optimal plant photochemical information pre-filter on Sentinel-2 bands. And then, the $VI_{LAI}$ and $VI_{LCC}$ time-series will used as the biological agents of the plant canopy structure and plant biochemical state in the follow analysis.   \par

\textit{b.Fast Fourier Convolutional Layer} \par
The input biological agent (i.e. $VI_{LAI}$ or $VI_{LCC}$) dynamics extracted from the Sentinel-2 time-series can be viewed as a sample patch with $k \times k$ pixel vectors. Each of the pixels represents a class with $K^{(1)}$ time-series channels. Then, the 3D patches with a size of $k \times k \times K^{(1)}$ are extracted as the input of the fast Fourier convolution layer. \par

The fast Fourier convolution is used to decomposes the biological agent time-series into a series of frequency components with various time-scales based on the fast Fourier transform. Mathematically, FFT decomposes the original time-series signal $f(t)$ to the frequency domain by the linear combination of trigonometric functions as follows:
\begin{equation}
F(\varpi)=\int_ { - \infty }^{ + \infty } {f(t) e^{-i \varpi t}dt}
\end{equation}
Where $\varpi$ is frequency and $ F(\varpi)$ is the Fourier coefficient with frequency $\varpi$, $i$ is the unit of imaginary number. It is customary to use a discrete form as follow: 
\begin{equation}
F(x)_{k \times k}=\frac{1}{K^{1}} \sum_{n=0}^{K^{1}-1} x_{n} e^{-\frac{2 \Pi xni}{K^{1}}}
\end{equation}
Where $x=0,1,2,…N-1$, and $N$ is the length of time series. \par

Among the frequency-domain components of the biological agents of $VI_{LAI}$ and $VI_{LCC}$ dynamics, the low-frequency components always indicate the soil background or phenological characteristics of the ground entities. The high-frequency region generally represents environmental noises, such as landcover variations or illumination inconsistency. Therefore, considering the infestation and development of yellow rust and nitrogen deficiency is a continuous biological process on the proxies of $VI_{LAI}$ and $VI_{LCC}$, we hypothesize that the medium-frequency region represents the yellow rust and nitrogen deficiency associated $VI_{LAI}$ and $VI_{LCC}$ fluctuations. Thus, the yellow rust and nitrogen deficiency associated responses can be characterized from the background and environmental noises by an optimized activation function. In this study, the ReLU activation function is implemented to calculate the $VI_{LAI}$ and $VI_{LCC}$ time-series magnitude in the medium-frequency region, and the activation feature map, with a size of $k \times k \times K^{(2)}$, is conducted with Fourier Pool layer to extracted the sensitive $VI_{LAI}$ and $VI_{LCC}$ response in frequency-domain, and output the fast Fourier convolution (FFC) features .\par 

\textit{c. Capsule Feature Encoder} \par
\begin{figure}[]   
    \centering  
    \includegraphics[width=5.5in]{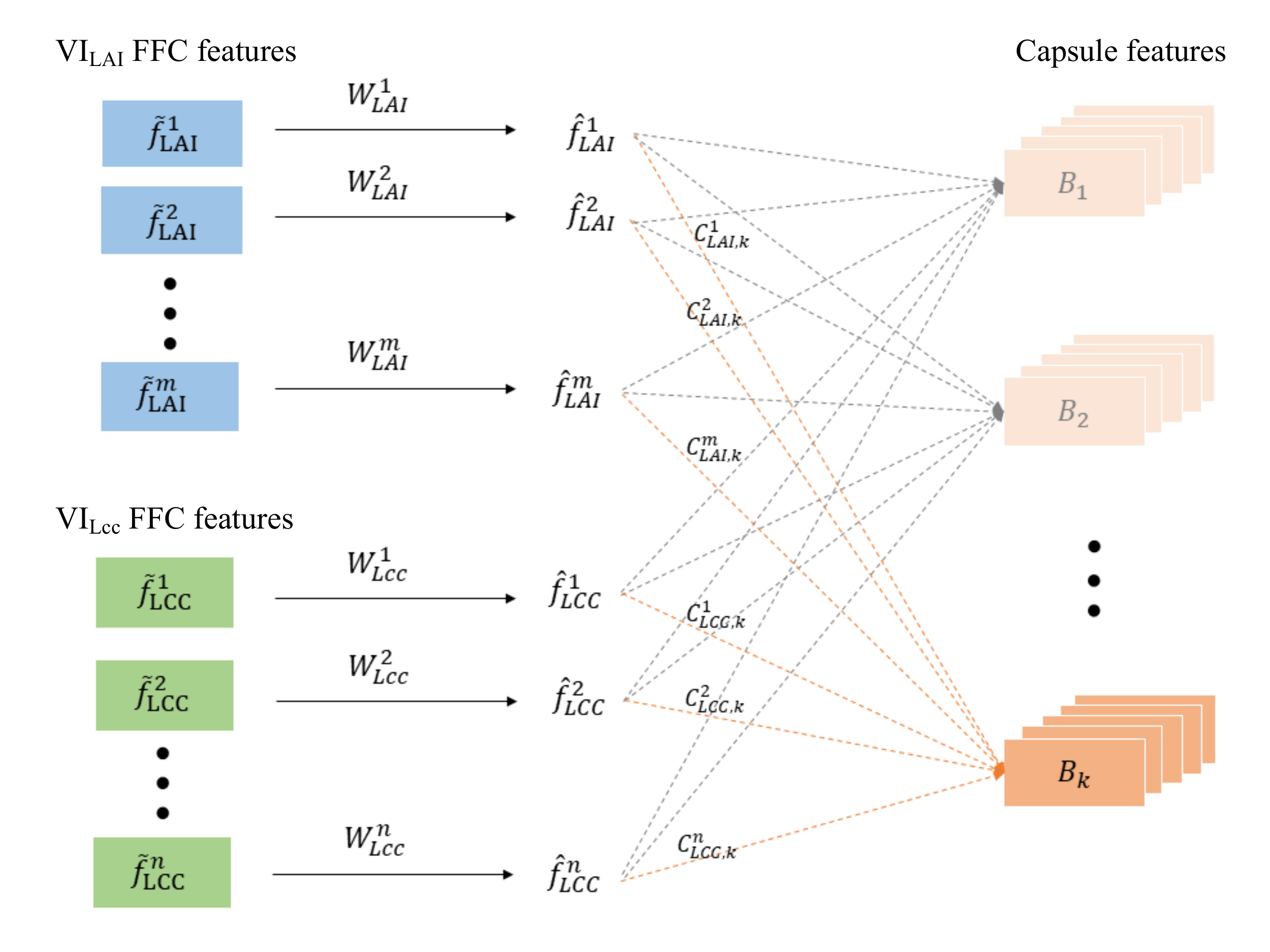}   
    \caption{ The dynamic routing optimization between the FFC scalar features and the capsule vector features}  
    \label{fig:7}  
\end{figure}

Considering the host-stress interaction of the plant stresses is a comprehensive progress that co-represented by the various biological agents. Therefore, modelling the part-to-whole relationship is the most significant evidences for detection and discrimination of plant stresses. We develop a capsule feature encoder to rearrange the extracted $VI_{LAI}$ and $VI_{LCC}$ FFC features, which are the scalar features, into the joint capsule vector features. Thees joint vector features represent the hierarchical structure of the $VI_{LAI}$ and $VI_{LCC}$ responses to the specific plant stress. It’s noteworthy that the extracted $VI_{LAI}$ and $VI_{LCC}$ scalar FFC features themselves respectively represent the biophysical and biochemical response to the plant stress development. Therefore, the joint vector features have great performance to characterize the intrinsic entanglement of host-stress interaction. In order to optimize the learning process between the the FFC scalar features and the capsule vector features, and dynamic routing algorithm is introduced as shown in Fig. \ref{fig:7}.

Specifically, the $VI_{LAI}$ and $VI_{LCC}$ FFC features, $\{\bar{f}_{LAI}^{1},  \bar{f}_{LAI}^{2},…, \bar{f}_{LAI}^{m}, \bar{f}_{LCC}^{1}, \bar{f}_{LCC}^{2},…, \bar{f}_{LCC}^{n}\}$, are firstly normalized by using the normalization weights $W \in \{W_{LAI}^1, W_{LAI}^2,…, W_{LAI}^m, W_{LCC}^1, W_{LCC}^2,…, W_{LCC}^n\}$. 
This step smooths the feature values and makes them obey a normal distribution. In addition, this normalization operation is helpful for retraining the vanishing gradients in the back-propagation progress. After that, the normalized FFC features, $\{\hat{f}_{LAI}^{1},  \hat{f}_{LAI}^{2},…, \hat{f}_{LAI}^{m}, \hat{f}_{LCC}^{1},  \hat{f}_{LCC}^{2},…, \hat{f}_{LCC}^{n}\}$, are rearranged into $K^{3}$ capsule features with the coupling coefficients of $c$. Here, $c$ is a series of trainable parameters that encodes the part-whole relationships between the FFC scalar features and the capsule vector features. The translation and orientation of the capsule vector feature represents the class-specific hierarchical structure characteristics in terms of $VI_{LAI}$ and $VI_{LCC}$ responses in frequency-domain, while its length represents the degree a capsule is corresponding to a class. To measure the length of the output vector as a probability value, a nonlinear squash function is used as follow:

\begin{equation}
\breve u_m=\frac{||u_m||^2}{1+||u_m||^2}\cdot\frac{u_m}{||u_m||}
\end{equation}

wherein, $\breve u_m^{(l)}$ is the scaled vector of $\mathbf{X}_{out}^2$. This function compresses the short vector features to zero and enlarge the long vector features a value close to 1. The final output is denoted as $\mathbf{X}_{out}^3 \in \mathbb{R}^{ Z \times 1 \times 1 \times K}$.  \par

Finally, the $K^{3}$ capsule features will be weightily combined into $Z$ class capsules and the final outputs are the class-wised biological composed feature $=\{V_1, V_2, …, V_Z\}$. In this study, $Z$ is 3 because of three interested classes (i.e. healthy wheat, yellow rust, and nitrogen deficiency).

\textit{d. Classifier} \par
Based on the characteristics of the class-capsule feature vectors, a classifier is defined to achieve the final detection and discrimination. This classifier is composed by two layers: an activation layer and a classification layer. \par
Specifically, The activate function is defined as: 
\begin{equation}
\hat{V}_h=\frac{{\Vert V_h \Vert}^2}{1+{\Vert V_h \Vert}} \cdot \frac{V_h}{\Vert V_h \Vert}
\end{equation}
Where, $V_h$ is the class-capsule features corresponding to class $h \leqq Z$. $\Vert \cdot \Vert$ indicates the operator of 1-norm. In fact, the orientation of the $\hat{V}_h$ represents the instantiation parameters of the biological responses for the class $h$, and the length represents the membership that the feature belongs to class $h$. And then, an argmax function is used to achieve the final classification by seeking the largest length of $\hat{V}_h$. The argmax function is defined as: 
\begin{equation}
\mathop{argmax}\limits_{h} O_{i,j}^{5}=\{h | \forall g: \Vert V_g \Vert < \Vert V_h \Vert\}
\end{equation}

\section{Materials and experiments}

In this study, we use nitrogen deficiency and the yellow rust as the study cases for model testing and evaluation. In order to comprehensively test and evaluate the classification accuracy, robustness and generalization of the proposed model, we collected two types of the data: 1) the high-quality labelled dataset under the controlled field conditions, 2) the ground survey dataset under the natural field conditions. The former is used for training and optimizing the proposed model, the latter is used for testing and evaluating the generalization and transferability of the well-trained model in the actual application cases. The detailed information is described as below: \par

\subsection{Study sites} \par
To avoid the fungus contamination on the other groups, we respectively carried out two independent experiments under similar environmental conditions recording continuous in-situ observations of: a) yellow rust infestation from $20^{th}$ April to  $25^{th}$ May 2017 at the Scientific Research and Experimental Station of Chinese Academy of Agricultural Science ($39^{\circ} 30^{\prime} 40^{\prime \prime} N$, $116^{\circ} 36^{\prime} 20^{\prime \prime} E$) in Langfang, Hebei province and b) nitrogen deficiency at the National Experiment Station for Precision Agriculture ($40^{\circ} 10^{\prime} 6^{\prime \prime} N$, $116^{\circ} 26^{\prime} 3^{\prime \prime} E$) in Changping District, Beijing, China. The measurement strategies focused on eight key wheat growth stage (i.e. jointing stage, flag leaf stage, heading stage, flowering stage, early grain-filling stage, mid grain-filling stage, late grain-filling stage, and harvest stage). The detailed observation dates and the canopy photos were listed in Table 1. The same experiments were repeated from $18^{th}$ April to $31^{th}$ May, 2018.\par

For the yellow rust experiment, we used the wheat cultivar ‘Mingxian 169’ due to its susceptibility to yellow rust infestation. There was a control group and two infected groups of yellow rust (two replicates of inoculated treatment) were applied. Each field group occupied $220m^2$ of field campaigns in which there were 8 planting rows. For the control group, a total of 8 plots (one plot in each row) with an area of $1m^2$ were symmetrically selected in the field for hyperspectral observations and biophysical measurements. For the disease groups, the concentration levels of $5 mg 100^{-1} ml^{-1}$ and $9 mg 100^{-1} ml^{-1}$ spores solution was implemented to generate a gradient in infestation levels, 8 plots were applied for sampling in each replicate, respectively. All treatments applied $200 kg ha^{-1}$ nitrogen and $450 m^{3} ha^{-1}$ water at the beginning of planting. \par
For the nitrogen deficiency experiment in Changping, the popular wheat cultivars, ‘Jingdong 18’ and ‘Lunxuan 167’, were selected. There were two replicates field groups with same nitrogen treatment were applied. Each field group occupied $600 m^2$ of field campaigns in which three fertilization levels were used in 21 planting rows of field land (7 rows per treatment) at the beginning of planting, $0 kg \cdot ha^{-1}$  nitrogen (deficiency group), $100 kg \cdot ha^{-1}$ nitrogen (deficiency group), and $200 kg \cdot ha^{-1}$ nitrogen (control group). Similarly to Langfang, all treatments received $450 m^{3} \cdot ha^{-1}$  water at planting. 

\begin{center}
\setlength\LTleft{0pt}
\setlength\LTright{0pt}
\begin{longtable}{  c
  c
  m{3cm}
  m{3cm}
  m{3cm}
  m{3cm}
}
\caption{ The state of vegetation at each measurement dates.} \label{tab:1} \\

\toprule
Location\\(year) & Type & \multicolumn{4}{c}{Day after treatment (DAT)} \\
\midrule
\multicolumn{1}{c}{\multirow{6}{4em}{Langfang \\ 2017}} &  & \multicolumn{1}{c}{7(Apr.20)} & \multicolumn{1}{c}{14(Apr.27)} & \multicolumn{1}{c}{23 (May.6)} & \multicolumn{1}{c}{27(May.10)}\\
& \multicolumn{1}{c}{H} & \includegraphics[width=3cm,height=3cm]{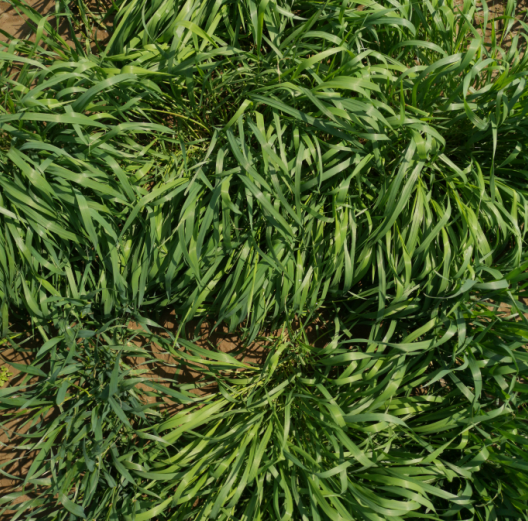} & \includegraphics[width=3cm,height=3cm]{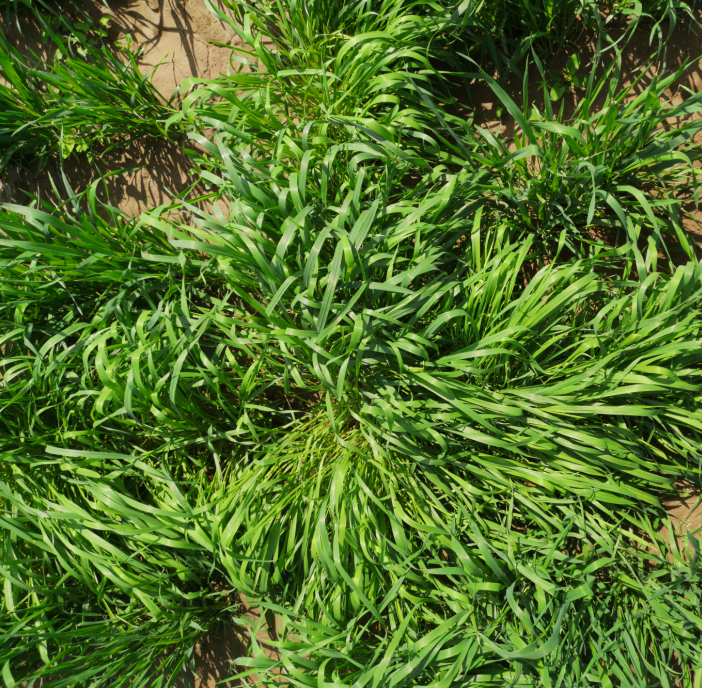} & \includegraphics[width=3cm]{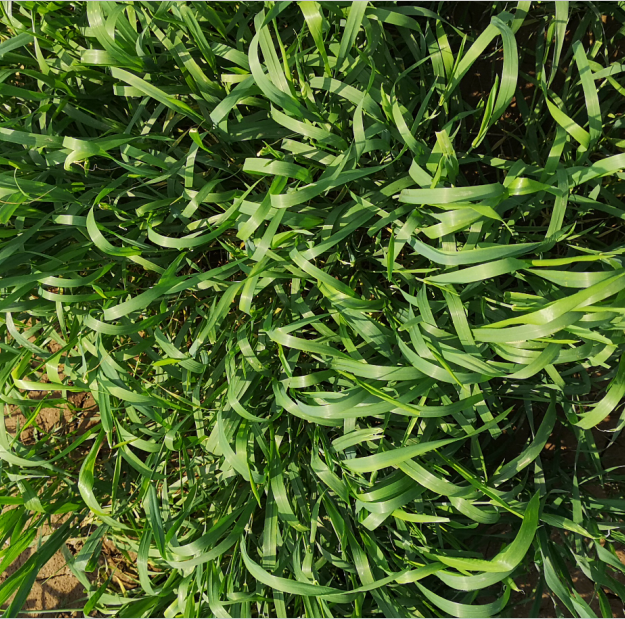} & \includegraphics[width=3cm]{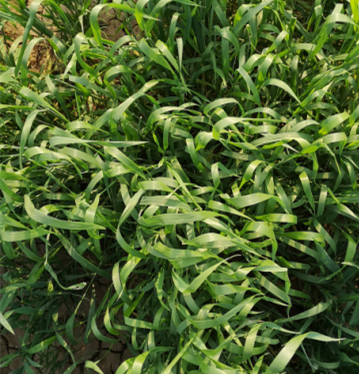} \\
& YR & \includegraphics[width=3cm,height=3cm]{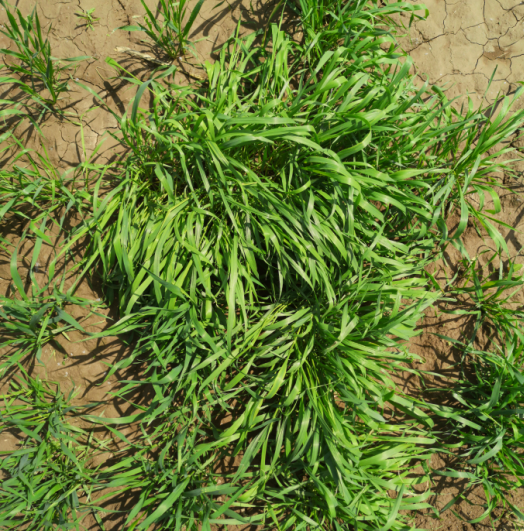} & \includegraphics[width=3cm,height=3cm]{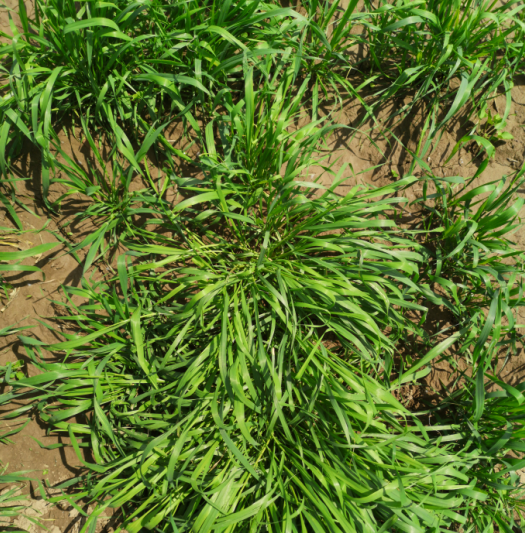} & \includegraphics[width=3cm,height=3cm]{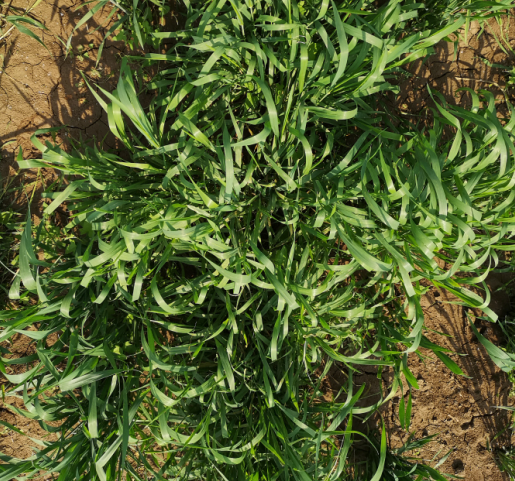} & \includegraphics[width=3cm,height=3cm]{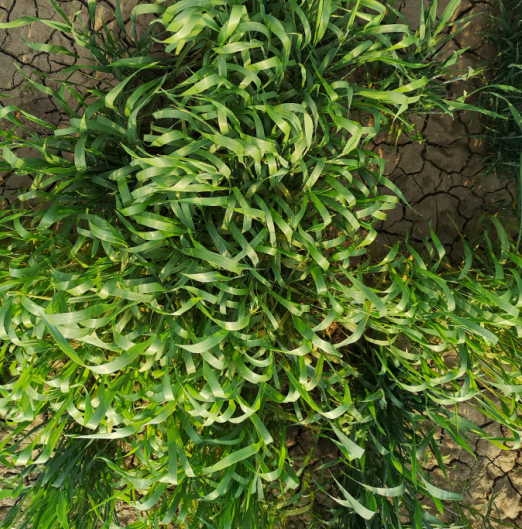} \\
\cmidrule{2-6}
&  & \multicolumn{1}{c}{34(May.17)} & \multicolumn{1}{c}{37(May.20)} & \multicolumn{1}{c}{41 (May.25)} & \\
& H & \includegraphics[width=3cm,height=3cm]{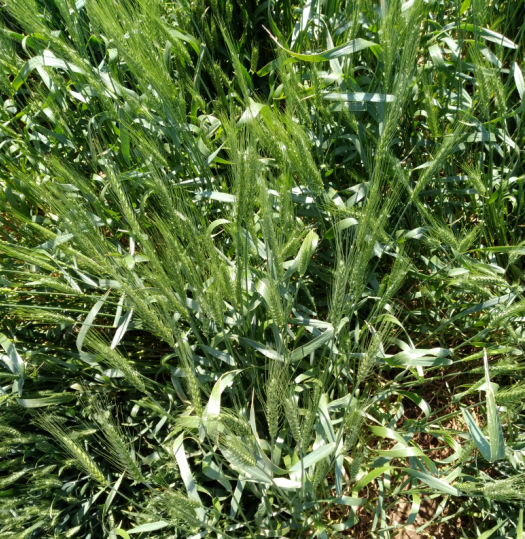} & \includegraphics[width=3cm,height=3cm]{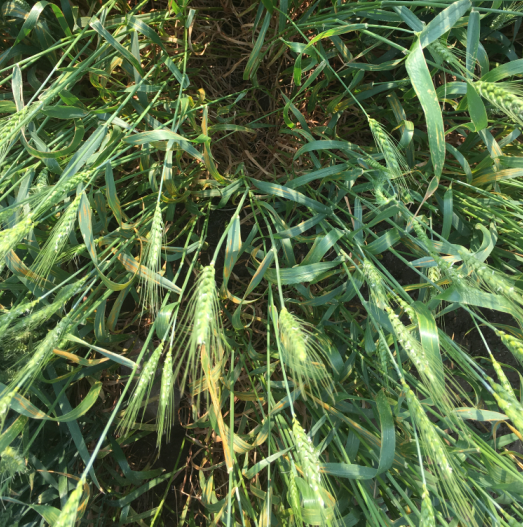} & \includegraphics[width=3cm,height=3cm]{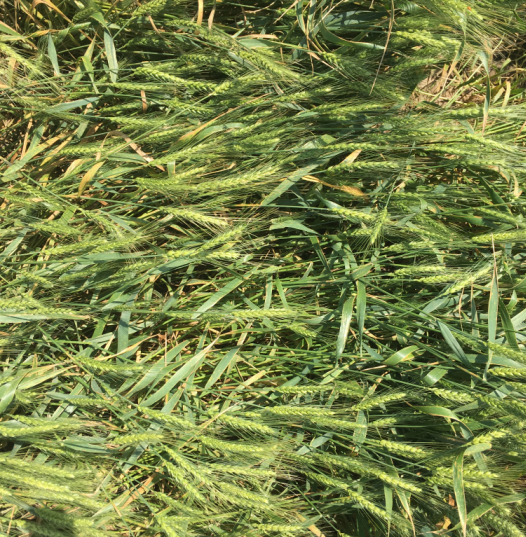} &  \\
& YR & \includegraphics[width=3cm,height=3cm]{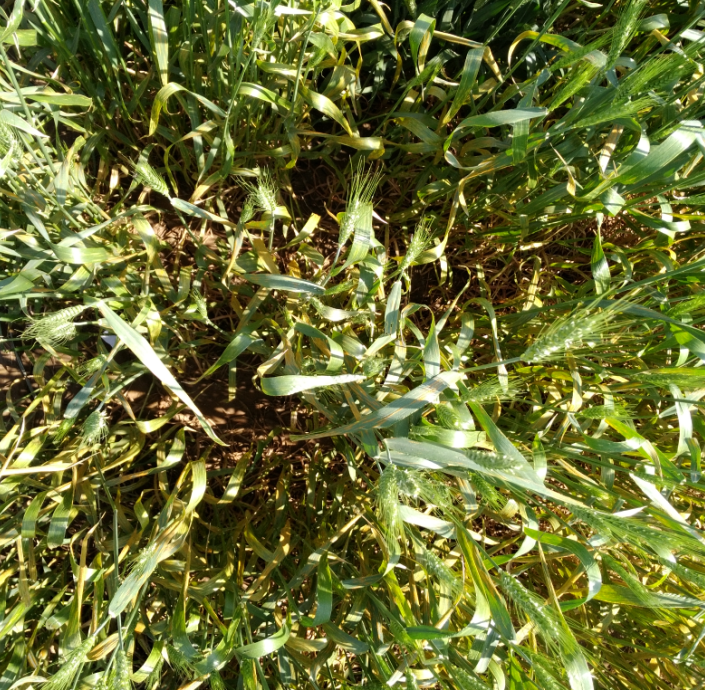} & \includegraphics[width=3cm,height=3cm]{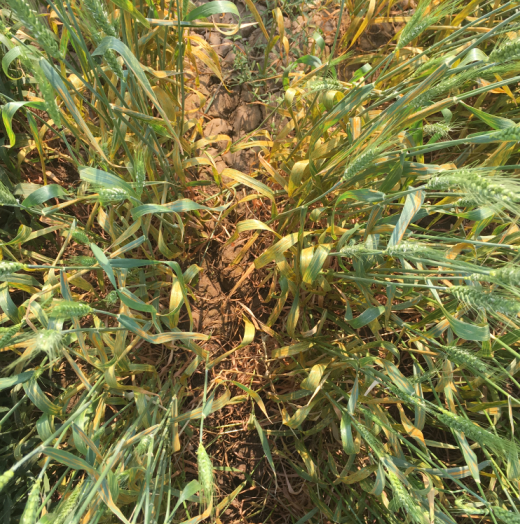} & \includegraphics[width=3cm,height=3cm]{table1/14.png} &  \\
\midrule
\multicolumn{1}{c}{\multirow{6}{4em}{Langfang \\ 2018}} &  & \multicolumn{1}{c}{7(Apr.18)} & \multicolumn{1}{c}{14(Apr.25)} & \multicolumn{1}{c}{23 (May.4)} & \multicolumn{1}{c}{27(May.8)}\\
& \multicolumn{1}{c}{H} & \includegraphics[width=3cm,height=3cm]{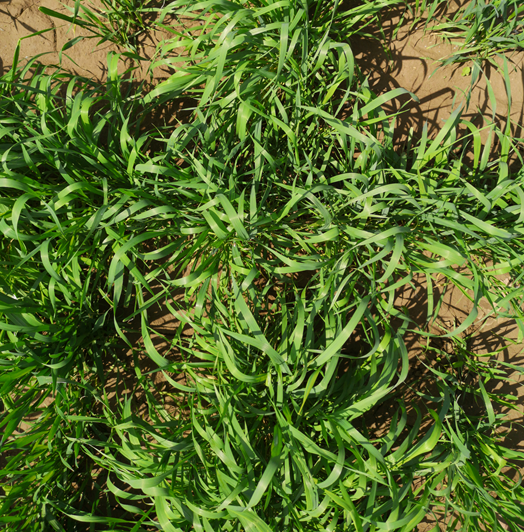} & \includegraphics[width=3cm,height=3cm]{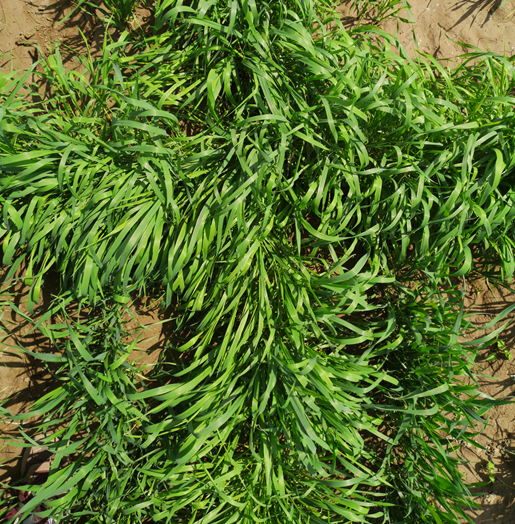} & \includegraphics[width=3cm]{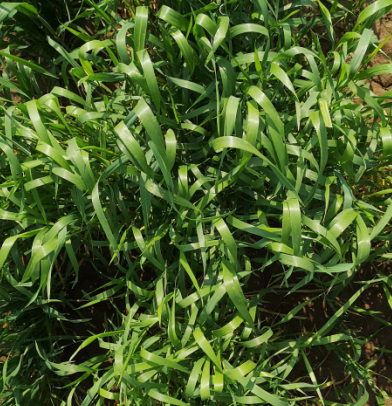} & \includegraphics[width=3cm]{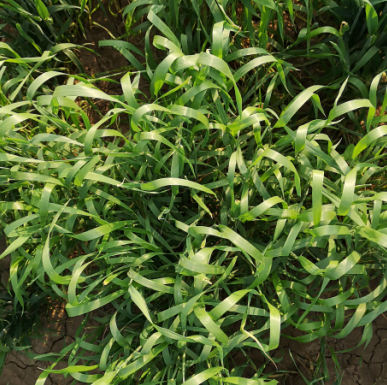} \\
& YR & \includegraphics[width=3cm,height=3cm]{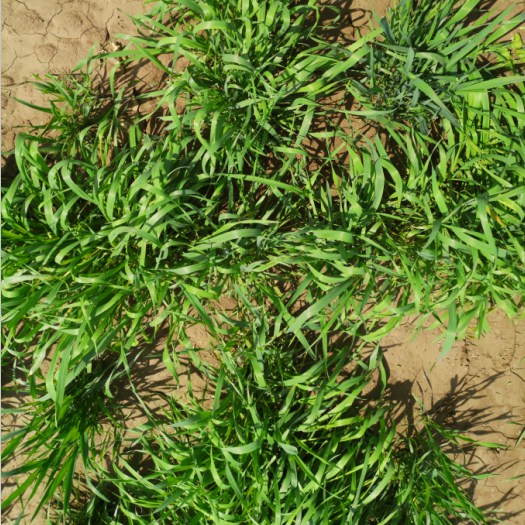} & \includegraphics[width=3cm,height=3cm]{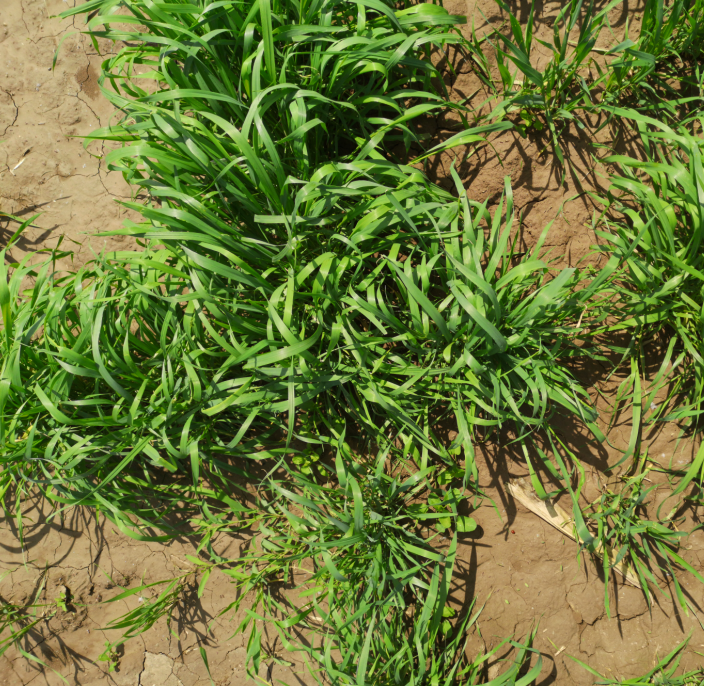} & \includegraphics[width=3cm,height=3cm]{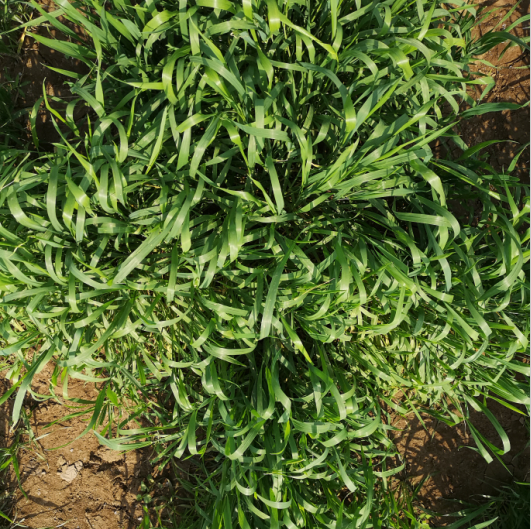} & \includegraphics[width=3cm,height=3cm]{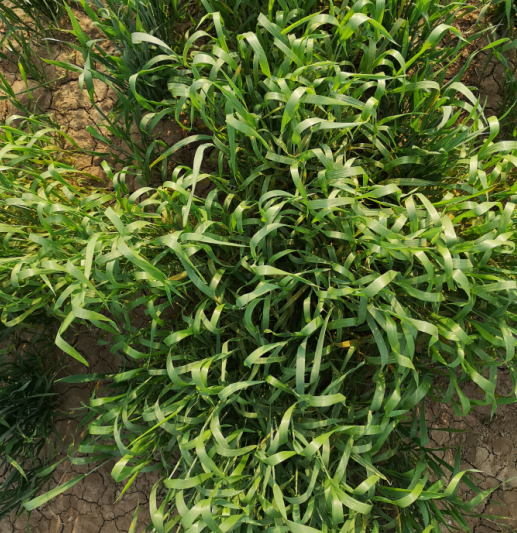} \\
\cmidrule{2-6}
&  & \multicolumn{1}{c}{34(May.15)} & \multicolumn{1}{c}{37(May.18)} & \multicolumn{1}{c}{41 (May.22)} & \multicolumn{1}{c}{49(May.30)} \\
& H & \includegraphics[width=3cm,height=3cm]{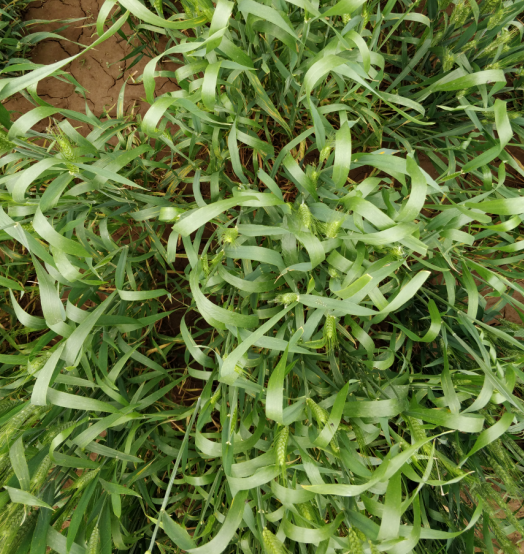} & \includegraphics[width=3cm,height=3cm]{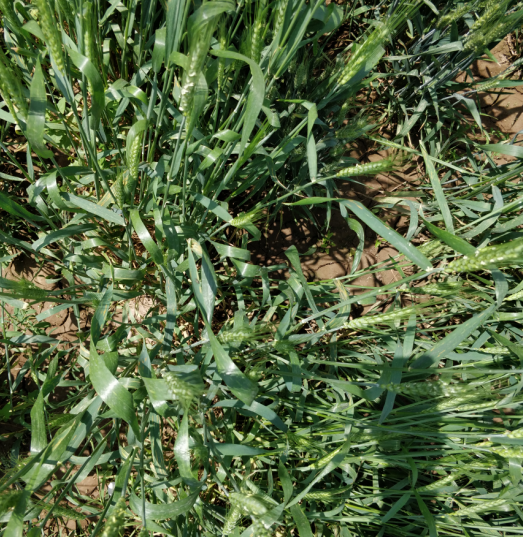} & \includegraphics[width=3cm,height=3cm]{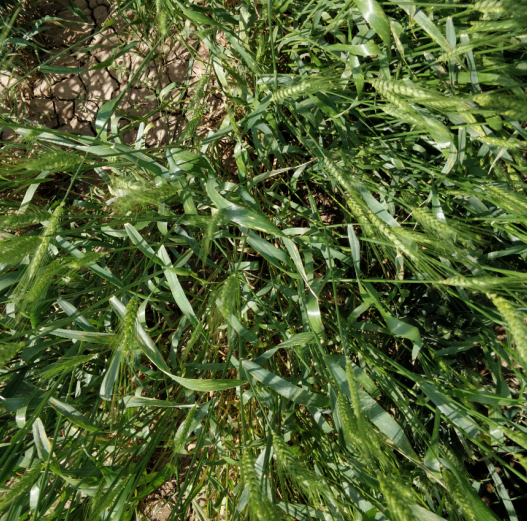} & \includegraphics[width=3cm,height=3cm]{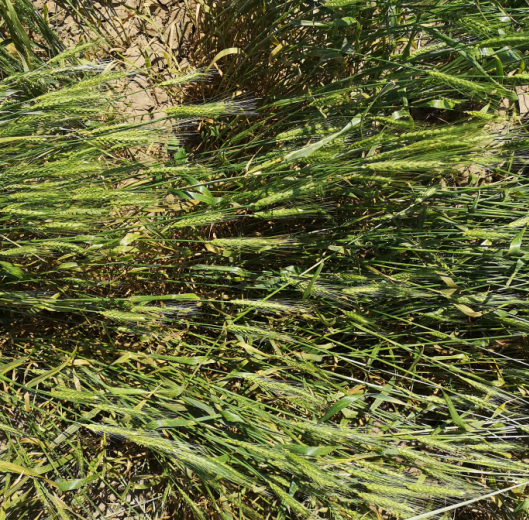} \\
& YR & \includegraphics[width=3cm,height=3cm]{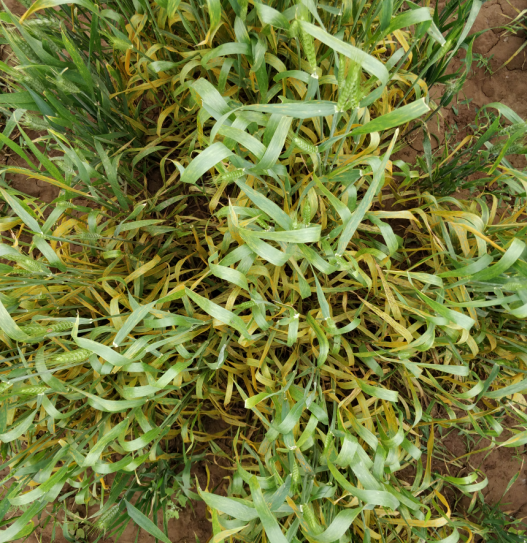} & \includegraphics[width=3cm,height=3cm]{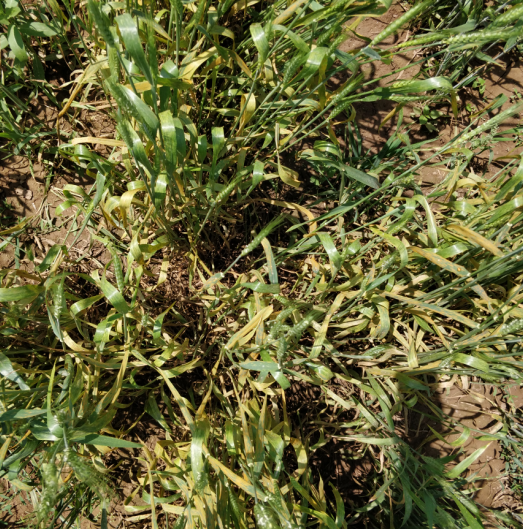} & \includegraphics[width=3cm,height=3cm]{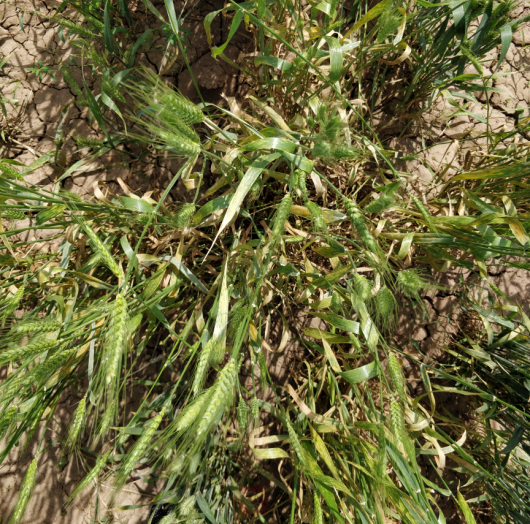} &\includegraphics[width=3cm,height=3cm]{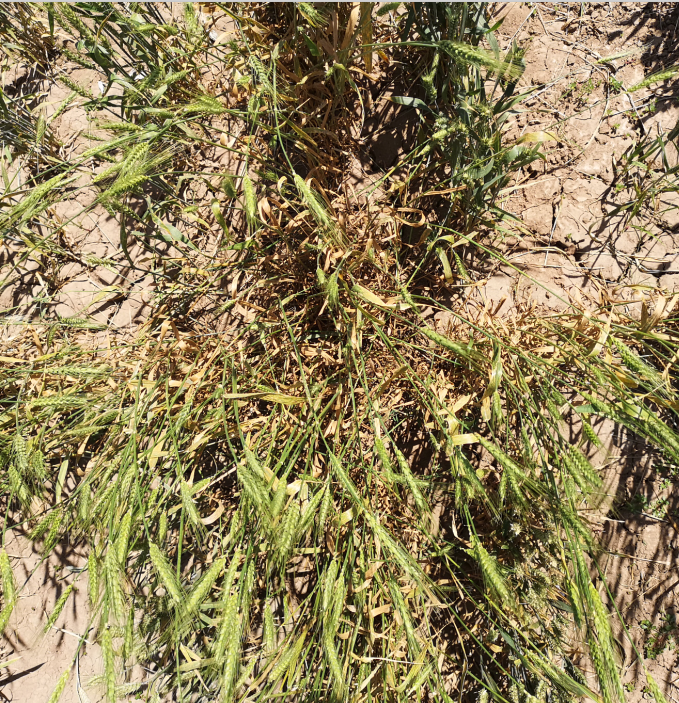}  \\
\midrule
\multicolumn{1}{c}{\multirow{6}{4em}{Xiaotang\\ shan \\ 2017}} &  & \multicolumn{1}{c}{7(Apr.16)} & \multicolumn{1}{c}{23(May.2)} & \multicolumn{1}{c}{34 (May.13)} & \multicolumn{1}{c}{49(May.29)}\\
& \multicolumn{1}{c}{H} & \includegraphics[width=3cm,height=3cm]{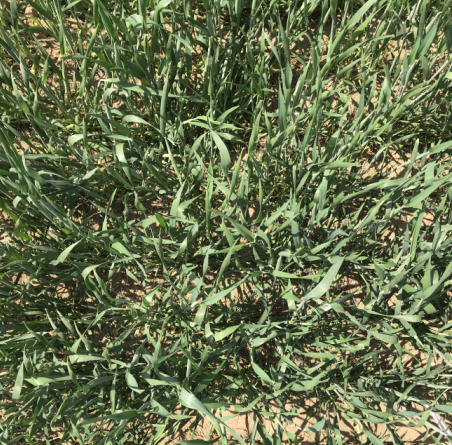} & \includegraphics[width=3cm,height=3cm]{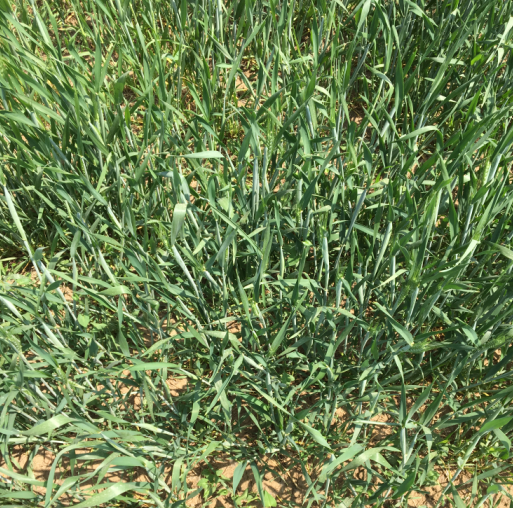} & \includegraphics[width=3cm]{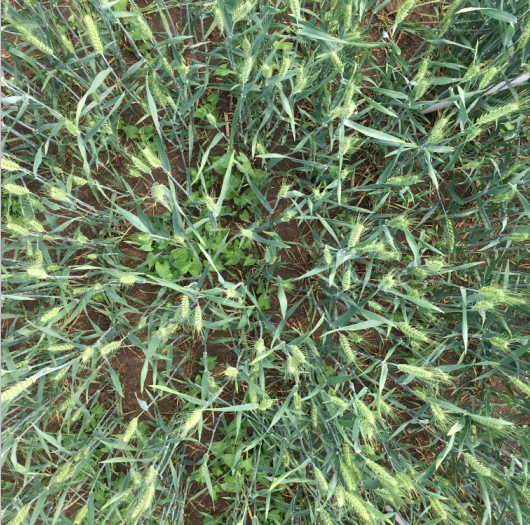} & \includegraphics[width=3cm]{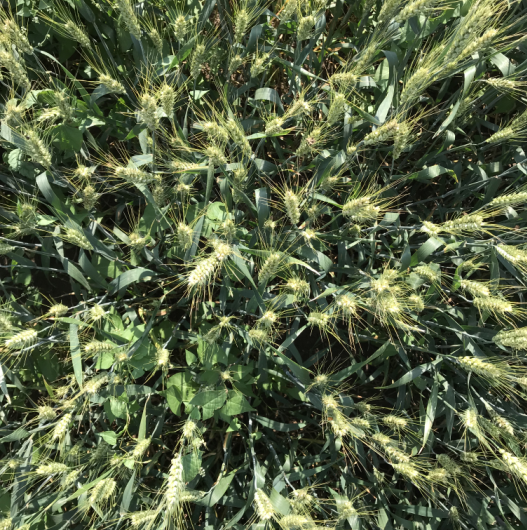} \\
& ND & \includegraphics[width=3cm,height=3cm]{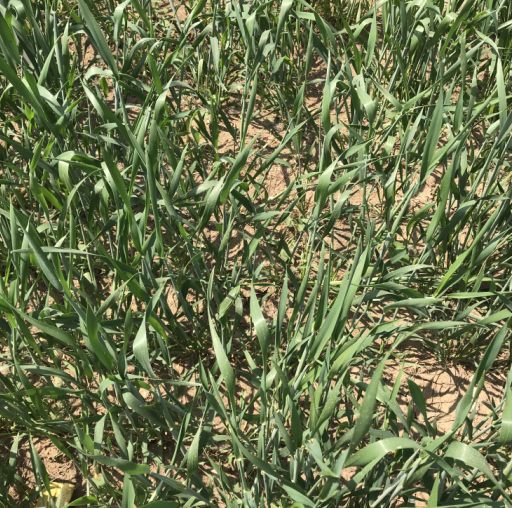} & \includegraphics[width=3cm,height=3cm]{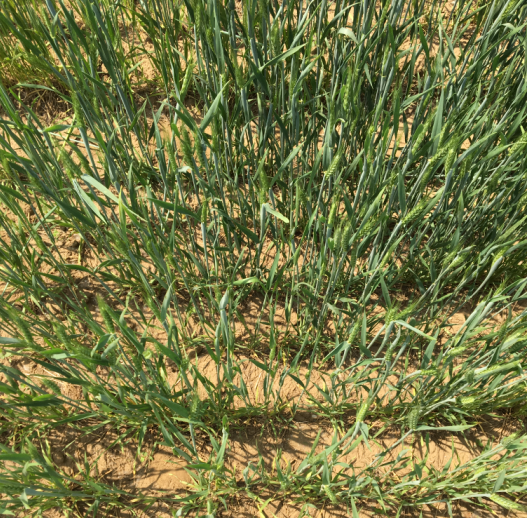} & \includegraphics[width=3cm,height=3cm]{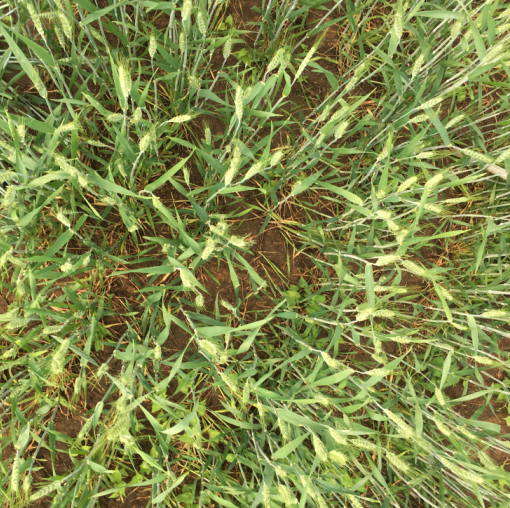} & \includegraphics[width=3cm,height=3cm]{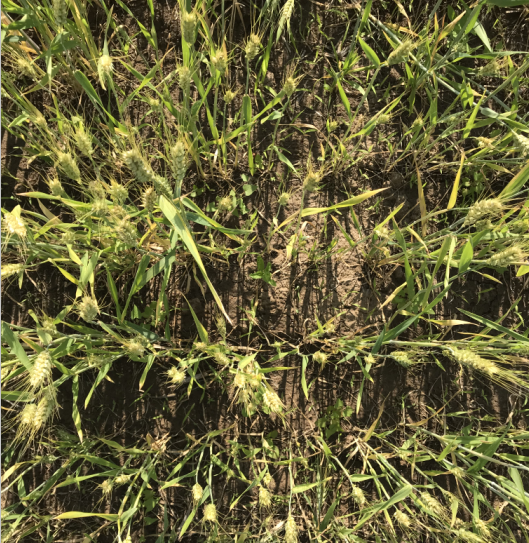} \\
\midrule
\multicolumn{1}{c}{\multirow{6}{4em}{Xiaotang\\ shan \\ 2018}} &  & \multicolumn{1}{c}{7(Apr.17)} & \multicolumn{1}{c}{23(May.5)} & \multicolumn{1}{c}{34 (May.14)} & \multicolumn{1}{c}{49(May.31)}\\
& \multicolumn{1}{c}{H} & \includegraphics[width=3cm,height=3cm]{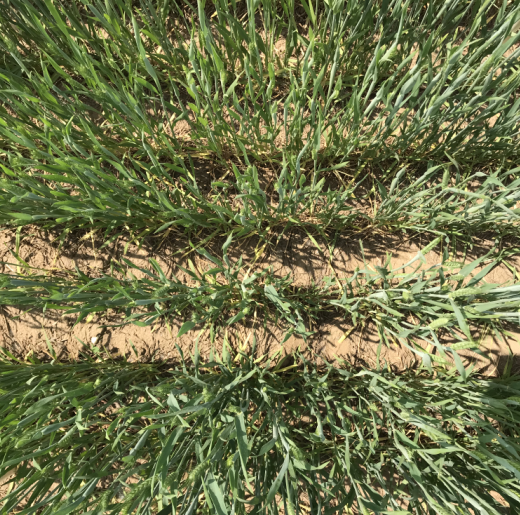} & \includegraphics[width=3cm,height=3cm]{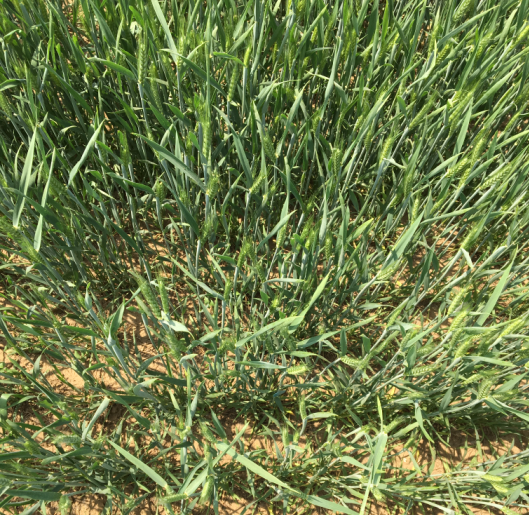} & \includegraphics[width=3cm]{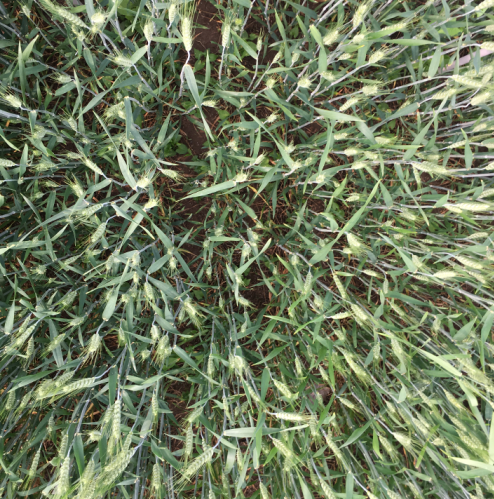} & \includegraphics[width=3cm]{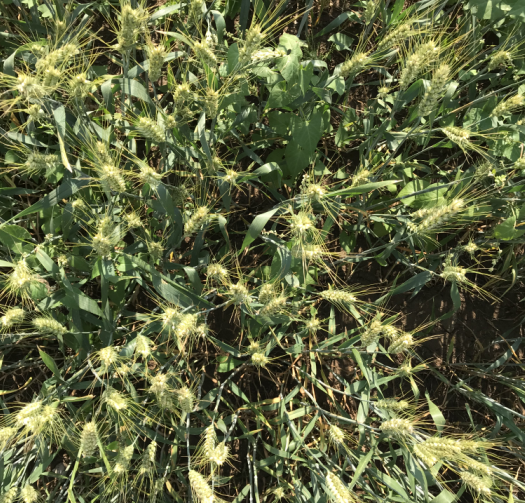} \\
& ND & \includegraphics[width=3cm,height=3cm]{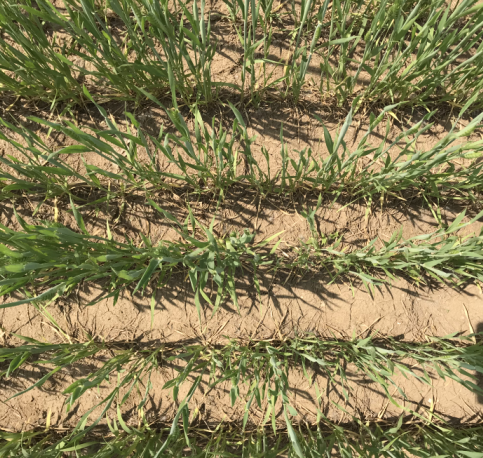} & \includegraphics[width=3cm,height=3cm]{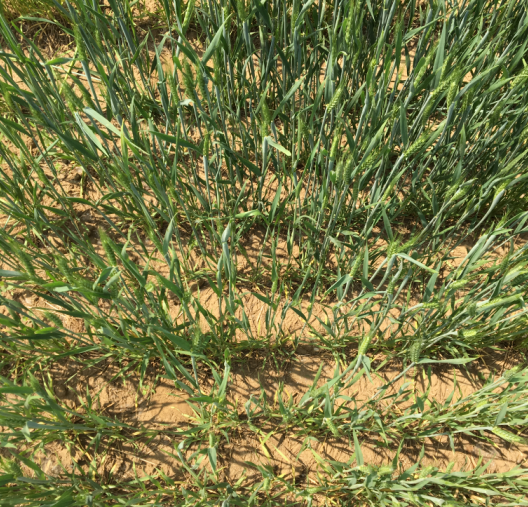} & \includegraphics[width=3cm,height=3cm]{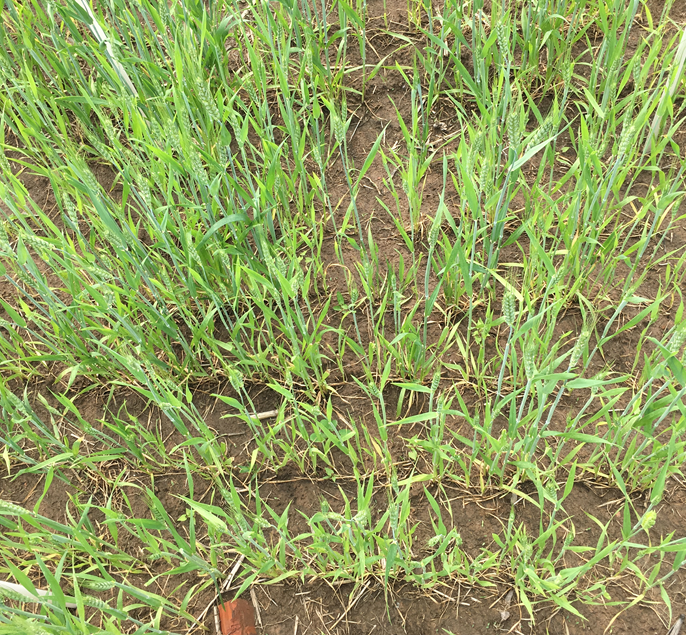} & \includegraphics[width=3cm,height=3cm]{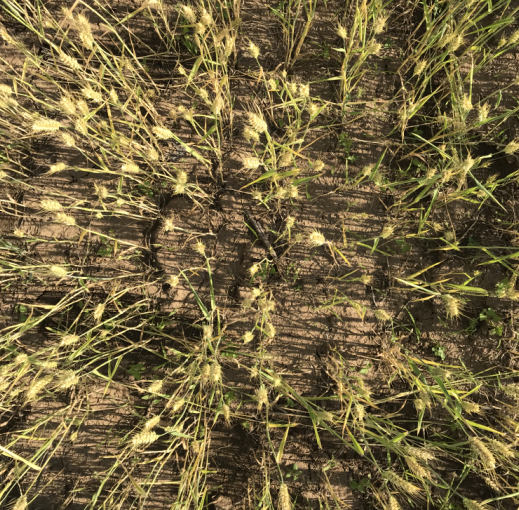} \\
\bottomrule

\end{longtable}
\end{center}
\noindent Note: H=healthy YR=yellow rust, ND=nitrogen deficiency \par

\subsection{The simulation of Sentinel-2 bands} \par

The simulated Sentinel-2 bands are regarded as the pure spectral signatures without the effects of atmosphere conditions. For this purpose, the reflectance and transmittances of the sampling plots were firstly collected using an ASD FieldSpec spectroradiometer (Analytical Spectral Devices, Inc., Boulder, CO, USA). In each plot, 10 scans were taken at 1.2m above the wheat canopy. The spectroradiometer was fitted with a $25^{\circ}$ field-of-view bare fiber-optic cable, and operated in the $350-2,500 nm$ spectral region. The sampling interval was $1.4 nm$ between $350$ and $1,050 nm$, and $2 nm$ between $1,050$ and $2,500 nm$. A white spectral reference panel ($99\%$ reflectance) was acquired once every 10 measurements to minimize the effect of possible difference in illumination. Only the bands in the range of $400–1000 nm$ were adopted in this study in order to match the visible-red edge-near infrared bands of Sentinel-2 and avoid bands below $400nm$ and above $1000 nm$ that were affected by noises \citep{RN27}. In order to keep radiance consistence, the sampling was conducted at the same period of time between 11:00 and 13:30 local time under a cloud-free sky. \par

Subsequently, we integrated the field canopy hyperspectral data with the sensor’s relative spectral response (RSR) function to simulate the multispectral bands of Sentinel-2. The formula is given as:
\begin{equation}
R_{sentinel-2}=\frac{\int_{\lambda_{start}}^{\lambda_{end}}{R_{ground}(\lambda) \cdot RSR(\lambda)\mathrm{d}x}}{\int_{\lambda_{start}}^{\lambda_{end}}{RSR(\lambda)\mathrm{d}x}}
\end{equation}
where $R_{sentinel-2}$ is the simulated multispectral channel of Sentinel-2 sensor, $\lambda_{start}$ and $\lambda_{end}$ represent the beginning and ending reflectance wavelength of Sentinel-2’s corresponding channel, respectively, and $R_{ground}$ is the ground truth canopy hyperspectral data, RSR is the relative spectral response of Sentinel-2 sensor (https://earth.esa.int/web/sentinel/user-guides/sentinel-2-msi/document-library/), both of the $R_{ground}$ and RSR are the functions of wavelength.\par

\subsection{Ground truth plant parameters collection}\par
The plant LAI and LCC were synchronously measured on the same place where the canopy spectral measurements were made. The LCC was measured by the Dualex Scientific sensor (FORCE-A, Inc. Orsay, France), a hand-held leaf-clip sensor designed to non-destructively evaluate the content of chlorophyll and epidermal flavonols. The LCC values were collected with the default unit, which were used preferentially because of the strong relationship between their digital readings and real foliar chlorophyll. Considering the canopy structure-derived multiple scattering process, the first three leaves from the top are regarded as the most effective one with maximum photosynthetic absorption rate, which not only represent the average growth state of the whole plant, but also contribute most to the canopy reflected radiation measured by our observations. Therefore, for each sampling plot, the first, second and third wheat leaves, from the top of ten randomly selected plant (30 leaves for each plot), were chosen for LCC measurements. For the LAI acquisition, the LAI-2200 Plant canopy analyzer (Li-Cor Biosciences Inc., Lincoln, NE, USA) was used in each $1 m \times 1 m$ subplot.\par

\subsection{Ground truth plant stress severity assessments} \par
\begin{figure}[]   
    \centering  
    \includegraphics[width=\textwidth]{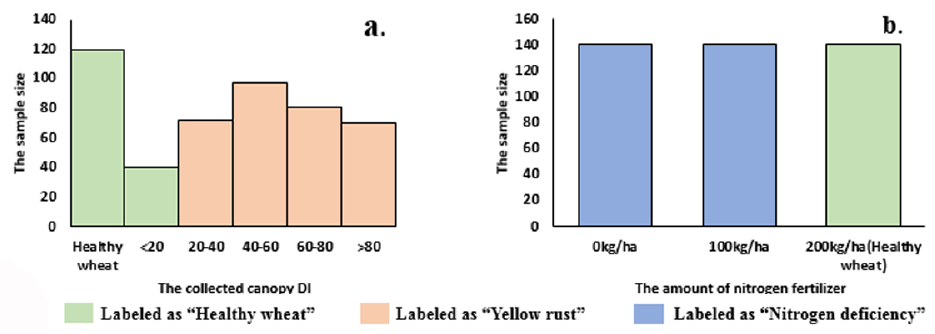}  
    \caption{ The distribution of the (a) collected disease index (DI) of yellow rust and (b) fertilization levels of nitrogen deficiency }  
    \label{fig:1}  
\end{figure}

In this study, the disease index (DI) was used to measure the severity of yellow rust, and the fertilization level was used to measure the severity of nitrogen deficiency. Specifically, the disease index (DI) was calculated using the method mentioned in \cite{RN10}. It’s noted that, because the slight stressed ($DI<20$) generate invisible influence on wheat yield and do not trigger enough spectral responses on the top of canopy (TOC) reflections of the $10m \times 10m$ Sentinel-2 pixels, the samples with $DI <20$ were labelled as “healthy wheat”, otherwise they were labelled as “yellow rust”. In order to guarantee the uniformed bias in each observation, all leaves were manual inspected by the same specially-assigned investigators according to the National Rules for the Investigation and Forecasting of Plant Diseases (GB/T 15795-1995). For nitrogen deficiency, three fertilization levels (i.e. $ 0 kg ha^{-1}$, $ 100 kg ha^{-1}$, and $ 200 kg ha^{-1}$) were controlled in our experiments, here we labelled the fertilization levels of$ 200 kg ha^{-1}$ as “healthy wheat“, otherwise they were labelled as “nitrogen deficiency”. The distribution of the collected DI of yellow rust and the fertilization levels of nitrogen deficiency are shown in Fig.\ref{fig:1}

\subsection{The ground survey dataset under the natural field conditions }
\begin{figure}[]   
    \centering  
    \includegraphics[width=\textwidth]{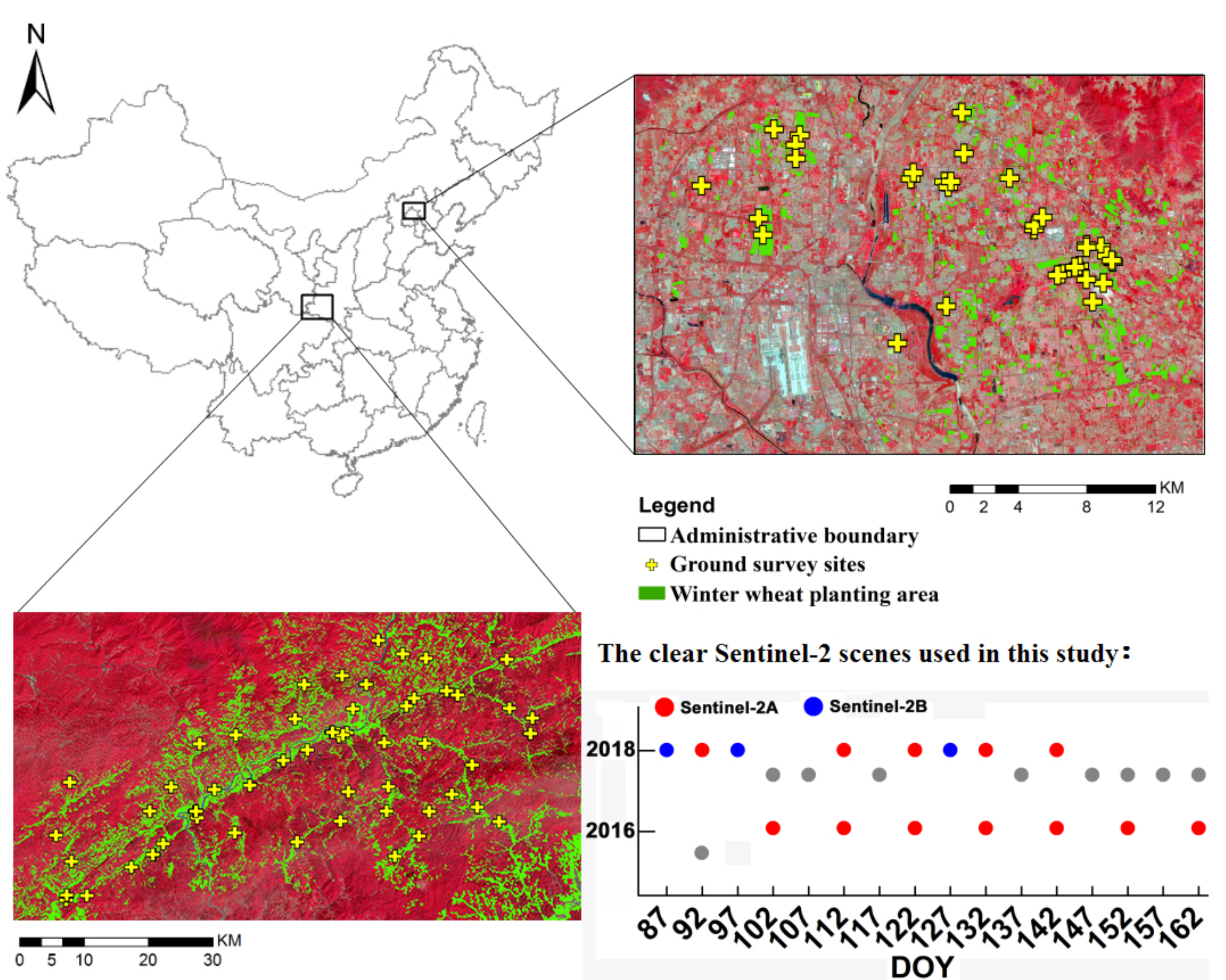}   
    \caption{ False-color maps of the experimental sites of Ningqiang county, Shaanxi, (bottom left) and Shunyi district, Beijing (top right). And the overview of the sentinel-2 imagery used.} 
    \label{fig:2}  
\end{figure}

In order to evaluate the generalization and transferability of the proposed model in the actual applications under the natural conditions, we collected the actual Sentinel-2 time-series and the ground truth data in two different sites, the one is located in the Ningqiang county ($37^{\circ} 35^{\prime} 51^{\prime \prime} N $, $118^{\circ} 35^{\prime} 19^{\prime \prime} E$), Shaanxi province, 2018, and another one is located in Shunyi district (($41^{\circ} 20^{\prime} 41^{\prime \prime} N $, $116^{\circ} 24^{\prime} 8^{\prime \prime} E$), Beijing, 2016. In Ningqiang county, a total of 9 cloud-free Sentinel-2 images and 55 ground truth plots were collected. In Shunyi district, a total of 6 cloud-free Sentinel-2 images and 32 ground truth were collected. All of the collected Sentinel-2 images were atmospherically corrected using the SEN2COR procedure, converting top-of-atmosphere (TOA) reflectance into top-of-canopy (TOC) reflectance. TOC products were the result of a resampling procedure with a constant ground resampling distance of 10 m for visible and near-infrared bands (B2, B3, B4, and B8) and 20m for red-edge bands (B5, B6, B7). The spatial resolution of the red-edge bands (B5, B6, B7) was homogenized to 10m using nearest neighbour resampling. Such process was conducted in the ESA SNAP 6.0 software. The basic principle of the nearest neighbour resampling was described in \cite{RN49}’s study. The overview of the sampling plots and Sentinel-2 collection are shown in Fig.\ref{fig:2}. 

\begin{figure}[]   
    \centering  
    \includegraphics[width=\textwidth]{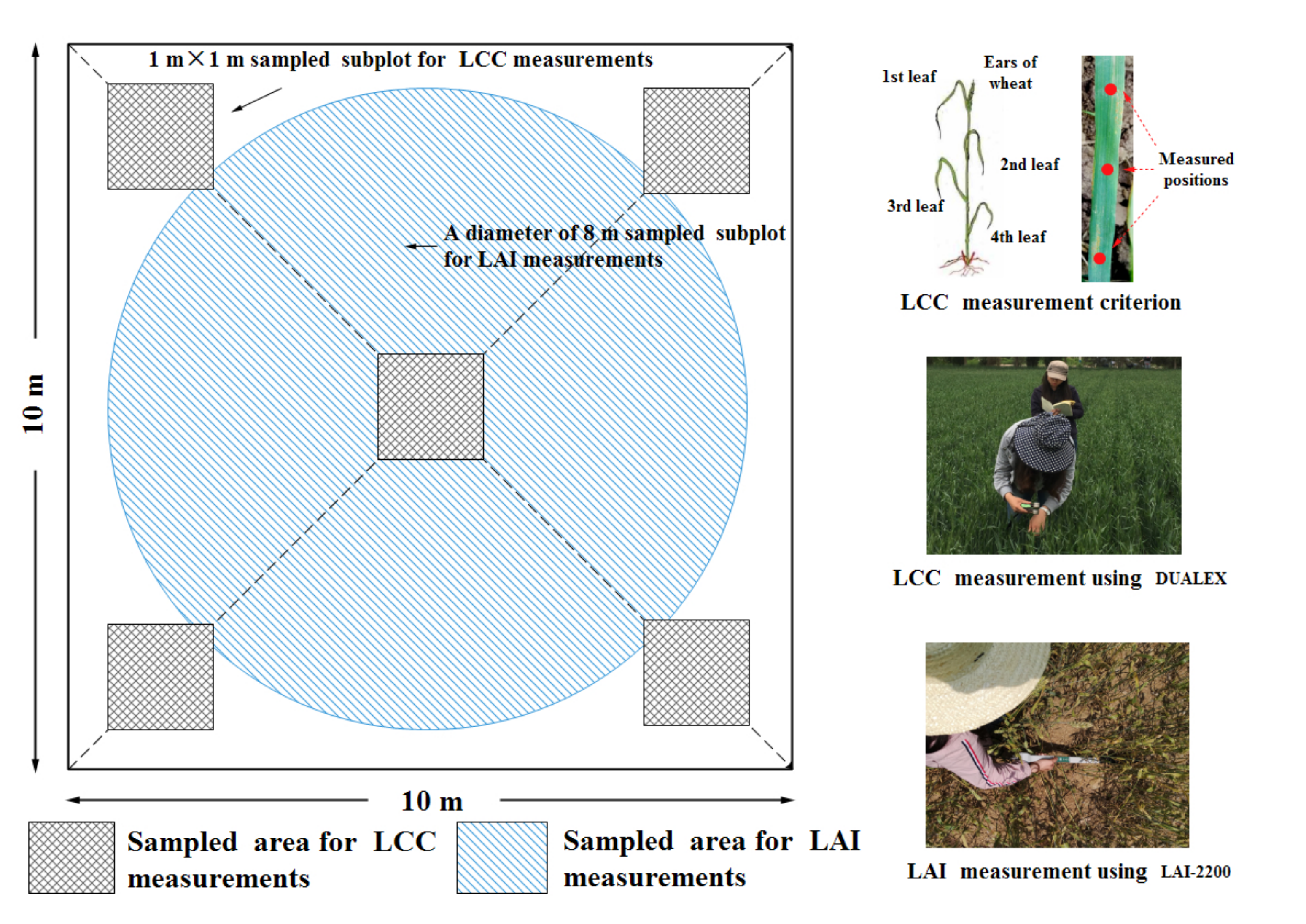}   
    \caption{ The measurement sketch of the synchronously ground LAI and LCC truth data collection.}  
    \label{fig:3}  
\end{figure}

In both surveys, LAI and LCC values were measured by the same approaches used in the experiments under controlled filed conditions. Each sample was collected in an area of about $10 m \times 10 m$ (for corresponding to the spatial resolution of Sentinel-2 bands), of which the centre coordinates were recorded using a GPS with differential correction (accuracy in the order of 2-5 m). The sketch of the sampled site setting is shown in Fig.\ref{fig:3}.

\begin{figure}[]   
    \centering  
    \includegraphics[width=\textwidth]{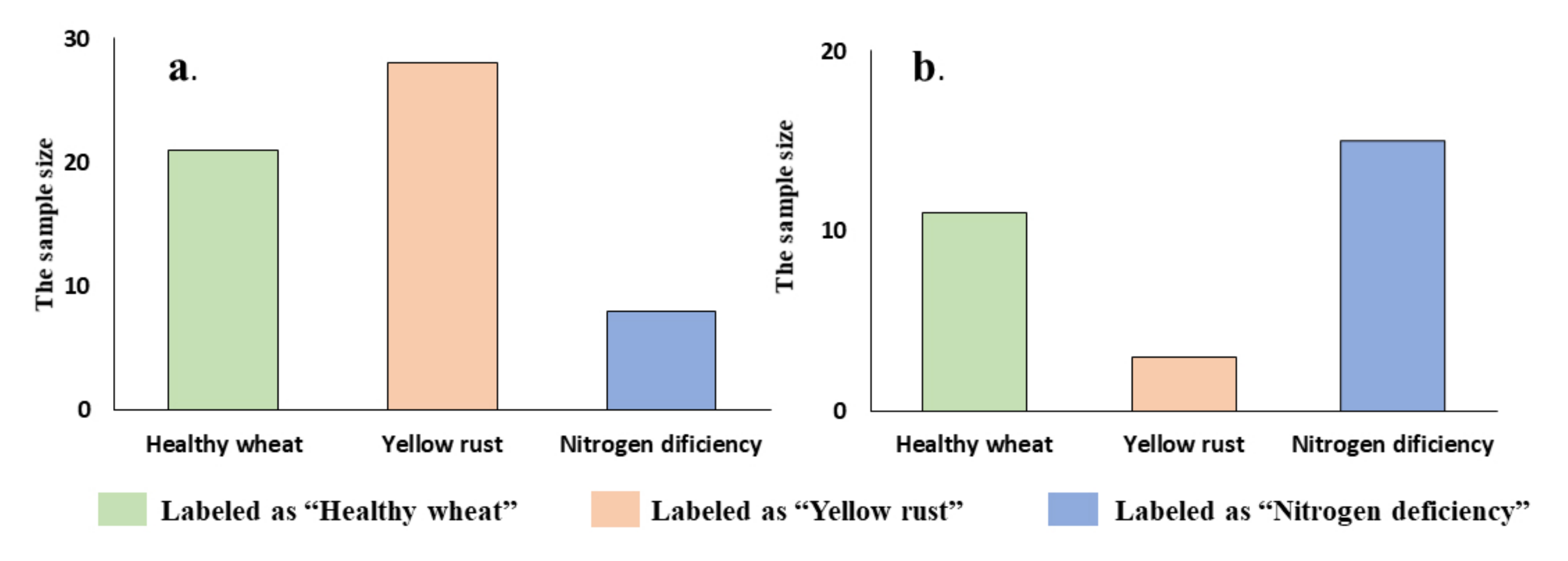}   
    \caption{ The distribution of the labeled classes in (a) Ningqiang and (b) Shunyi.}  
    \label{fig:4}  
\end{figure}

DIs of yellow rust were measured by the same method used in the experiments under controlled field conditions. In each plot, a plot was labeled as “yellow rust” when $DI>20$. On the other hand, nitrogen deficiency in each plot was investigated by requesting the history of fertilizer application to the local farmers, a plot was labeled as “nitrogen deficiency” when the history of fertilizer application $< 150 Kg/ha$. The statistical distribution of the labeled classes were shown in Fig.\ref{fig:4}.

\section{Results and discussion}
In this section, the proposed model is tested and evaluated in three different aspects, including the model performance on detecting and discriminating the yellow rust and nitrogen deficiency, computing efficiency, and the interpretability assessment.\par
Firstly, to test the performance of the proposed FFCDNN on detection and discrimination of yellow rust and nitrogen deficiency, two popular classification methods, thus, support vector machine (SVM) and convolutional neural network (CNN), are selected for experimental comparison and validation. Specifically, for the configuration of SVM classifier, the radial basis function (RBF) kernel is used in the SVM classification frame, and a grid-based approach proposed by \cite{RN55} is used to specify the parameter $C$ and $\delta$. For the CNN classifier, a network architecture with four convolutional layers and two fully connect layers proposed by \cite{RN80} is employed, all the hyperparameters of the CNN classifier have been optimized for the experiments. Regarding the model assessments, five widely used metrics, thus the average accuracy, producer’s accuracy, user’s accuracy, Kappa value, and computing time, are employed in this study to evaluate the classification accuracy. The definition of these matrices can be find in \citep{RN40} \par
Secondly, for the interpretability assessment of the model, a post-hoc analysis is used to expose the learning process and feature representations of the data life in the proposed model. Specifically, a canonical discriminant analysis is first used to measure the intra-class distance and the separability in each learning stage of the model. The definition of the canonical discriminant analysis is described in our previous study \citep{RN12}. And then, the coefficients of determination ($R^2$) between the generated biological composed features and the ground-measured severity of yellow rust and nitrogen deficiency are calculated based on univariate correlation analysis. 

\subsection{Model test on detection and discrimination of the yellow rust and nitrogen deficiency}
\subsubsection{Experiment one: model testing on the simulated Sentinel-2 bands under controlled filed conditions}
The first experiment is to evaluate the performance of the proposed model on the detection and discrimination of yellow rust and nitrogen based on the simulated Sentinel-2 bands under controlled field conditions. For model training and validation, a 5-folder cross validation is employed to measure the classification accuracy, and the computing time (CT) is used to measure the conputing effiency. 
The comparison of the classifications of the proposed FFCDNN, SVM, and CNN is shown in Table. \ref{tab:2}. Our results show that the proposed FFCDNN achieves the best classification in terms of accuracy and Kappa value on both of training dataset (with the overall accuracy of $92.8\%$ and Kappa of 0.891) and validation dataset (with the overall accuracy of $87.5\%$ and Kappa of 0.812). The misclassification mainly occurs between the healthy wheat and nitrogen deficiency. In the term of computing efficiency, althogh the computing time of the proposed model is not the best among the baseline, it is highly improved from the traditional convolution-basd deep learning model.   
\begin{center}
\setlength\LTleft{0pt}
\setlength\LTright{0pt}
\begin{longtable}{cccccccccccc}
\caption{ Accuracy evaluation for the classification of yellow rust and nitrogen based on the simulated Sentinel-2 bands under controlled field conditions .} \label{tab:2} \\
\toprule
&       & \multicolumn{5}{c}{Training}         & \multicolumn{5}{c}{Validation}      \\
\midrule
 &       & Health & YR  & NS   & U(\%) & OA(\%) & Health & YR   & NS & U(\%) & OA(\%) \\
\multirow{4}{*}{SVM} & Health & 160 & 10 & 12 & 87.9 & \multirow{4}{*}{88.3} & 97 & 8 & 15 & 80.8 & \multirow{4}{*}{79.7} \\
 & YR    & 6      & 169 & 8    & 92.3  &        & 7      & 106  & 13 & 84.1  &        \\
 & NS    & 14     & 13  & 148  & 84.6  &        & 16     & 14   & 84 & 73.7  &        \\
 & P(\%) & 88.9   & 88  & 88.1 &       &        & 80.8   & 82.8 & 75 &       &        \\
 & Kappa & \multicolumn{4}{c}{0.824}   &        & \multicolumn{4}{c}{0.695}  &        \\
 & CT(s) & \multicolumn{4}{c}{108.7}   &        & \multicolumn{4}{c}{32.5}   &        \\
\midrule
&       & \multicolumn{5}{c}{Training}         & \multicolumn{5}{c}{Validation}      \\
\midrule
 &       & Health & YR  & NS   & U(\%) & OA(\%) & Health & YR   & NS & U(\%) & OA(\%) \\
\multirow{4}{*}{CNN} & Health & 165 & 7 & 11 & 90.2 & \multirow{4}{*}{91.9} & 101 & 6 & 10 & 86.3 & \multirow{4}{*}{83.3} \\
 & YR    & 4      & 180 & 6    & 94.7  &        & 7      & 109  & 12 & 85.2  &        \\
 & NS    & 11     & 5  & 151  & 90.4  &        & 12     & 13   & 90 & 78.3  &        \\
 & P(\%) & 91.7   & 93.8  & 89.9 &       &        & 84.5   & 85.2 & 80.4 &       &        \\
 & Kappa & \multicolumn{4}{c}{0.877}   &        & \multicolumn{4}{c}{0.749}  &        \\
 & CT(s) & \multicolumn{4}{c}{481.5}   &        & \multicolumn{4}{c}{71.2}   &        \\
\midrule
&       & \multicolumn{5}{c}{Training}         & \multicolumn{5}{c}{Validation}      \\
\midrule
 &       & Health & YR  & NS   & U(\%) & OA(\%) & Health & YR   & NS & U(\%) & OA(\%) \\
\multirow{4}{*}{FFCDNN} & Health & 168 & 5 & 9 & 92.3 & \multirow{4}{*}{92.8} & 103 & 5 & 8 & 88.8 & \multirow{4}{*}{87.5} \\
 & YR    & 5      & 181 & 7    & 93.8  &        & 8      & 115  & 7 & 88.5  &        \\
 & NS    & 7     & 6  & 152  & 92.1  &        & 9     & 8   & 97 & 85.1  &        \\
 & P(\%) & 93.3   & 94.3  & 90.5 &       &        & 85.8   & 89.8 & 86.6 &       &        \\
 & Kappa & \multicolumn{4}{c}{0.891}   &        & \multicolumn{4}{c}{0.812}  &        \\
 & CT(s) & \multicolumn{4}{c}{277.4}   &        & \multicolumn{4}{c}{46.2}   &        \\
\bottomrule
\end{longtable}
\end{center}

\subsubsection{Experiment two: model test on the actual Sentinel-2 bands under natural filed conditions}

The second experiment aims to further evaluate the robustness and transferability of the proposed model on the actual Sentinel-2 images under natural field conditions. For this purpose, the pre-trained models in the last section are directly used in the pixel-wise classification of yellow rust and nitrogen deficiency on the actual Sentinel-2 time-series in Ningqiang county and Shunyi district, and the ground truth samples are used as validation. The accuracy assessment of the pre-trained SVM, CNN, and FFCDNN are listed in Table. \ref{tab:3}. Our results illustrate that the well-trained FFCDNN achieves an overall accuracy of $80.7\%$ (Kappa = 0.69) for the model test in Ningqiang, and an overall accuracy of $79.3\%$ (Kappa = 0.644) for the model test in Shunyi district. These accuracies are basically consistent with the results reported in experiment one. In comparison, there are evident declines could be figured out in the classification performance of SVM and CNN model, the average decline in term of overall accuracy respectively reaches $52.7\%$ for SVM and $36.3\%$ for CNN. Overall, these results suggest that the proposed FFCDNN provides a more stable and robust performance than the comprtitors on the classification of yellow rust and nitrogen deficiency in different field conditions.
\begin{center}
\setlength\LTleft{0pt}
\setlength\LTright{0pt}
\begin{longtable}{cccccccccccc}
\caption{ The accuracy assessment of the pre-trained models on actual Sentinel-2 time-series on classification of yellow rust and nitrogen deficiency in Ningqiang county of Shaanxi and Shunyi distinct of Beijing.} \label{tab:3} \\
\toprule
&       & \multicolumn{5}{c}{Ningqiang}         & \multicolumn{5}{c}{Shunyi}      \\
\midrule
 &       & Health & YR  & NS   & U(\%) & OA(\%) & Health & YR   & NS & U(\%) & OA(\%) \\
\multirow{4}{*}{SVM} & Health & 9 & 8 & 4 & 42.9 & \multirow{4}{*}{45.6} & 4 & 4 & 5 & 40 & \multirow{4}{*}{37.9} \\
 & YR    & 6      & 14 & 1    & 66.7  &        & 3      & 1  & 4 & 12.5  &        \\
 & NS    & 6     & 6  & 3  & 20  &        & 4     & 1   & 6 & 54.5  &        \\
 & P(\%) & 42.9   & 50  & 37.5 &       &        & 36.4   & 33.3 & 40 &       &        \\
 & Kappa & \multicolumn{4}{c}{0.158}   &        & \multicolumn{4}{c}{0.036}  &        \\
 & CT(s) & \multicolumn{4}{c}{112.6}   &        & \multicolumn{4}{c}{42.5}   &        \\
\midrule
&       & \multicolumn{5}{c}{Ningqiang}         & \multicolumn{5}{c}{Shunyi}      \\
\midrule
 &       & Health & YR  & NS   & U(\%) & OA(\%) & Health & YR   & NS & U(\%) & OA(\%) \\
\multirow{4}{*}{CNN} & Health & 11 & 5 & 3 & 57.9 & \multirow{4}{*}{57.9} & 5 & 1 & 4 & 50 & \multirow{4}{*}{55.2} \\
 & YR    & 4      & 18 & 1    & 78.3  &        & 2      & 2  & 2 & 33.3  &        \\
 & NS    & 6     & 5  & 4  & 26.7  &        & 4     & 0   & 9 & 69.2  &        \\
 & P(\%) & 52.4   & 64.3  & 50 &       &        & 45.5   & 66.7 & 60 &       &        \\
 & Kappa & \multicolumn{4}{c}{0.344}   &        & \multicolumn{4}{c}{0.272}  &        \\
 & CT(s) & \multicolumn{4}{c}{548.1}   &        & \multicolumn{4}{c}{102.5}   &        \\
\midrule
&       & \multicolumn{5}{c}{Ningqiang}         & \multicolumn{5}{c}{Shunyi}      \\
\midrule
 &       & Health & YR  & NS   & U(\%) & OA(\%) & Health & YR   & NS & U(\%) & OA(\%) \\
\multirow{4}{*}{FFCDNN} & Health & 16 & 1 & 2 & 84.2 & \multirow{4}{*}{80.7} & 9 & 1 & 2 & 75 & \multirow{4}{*}{79.3} \\
 & YR    & 2      & 24 & 0    & 92.3  &        & 0      & 2  & 1 & 66.7  &        \\
 & NS    & 3     & 3  & 6  & 50  &        & 2     & 0   & 12 & 85.7  &        \\
 & P(\%) & 76.2   & 85.7  & 75 &       &        & 81.8   & 66.7 & 80 &       &        \\
 & Kappa & \multicolumn{4}{c}{0.69}   &        & \multicolumn{4}{c}{0.644}  &        \\
 & CT(s) & \multicolumn{4}{c}{268.7}   &        & \multicolumn{4}{c}{81.6}   &        \\
\bottomrule
\end{longtable}
\end{center}

\begin{figure}[]   
    \centering  
    \includegraphics[width=5.5in]{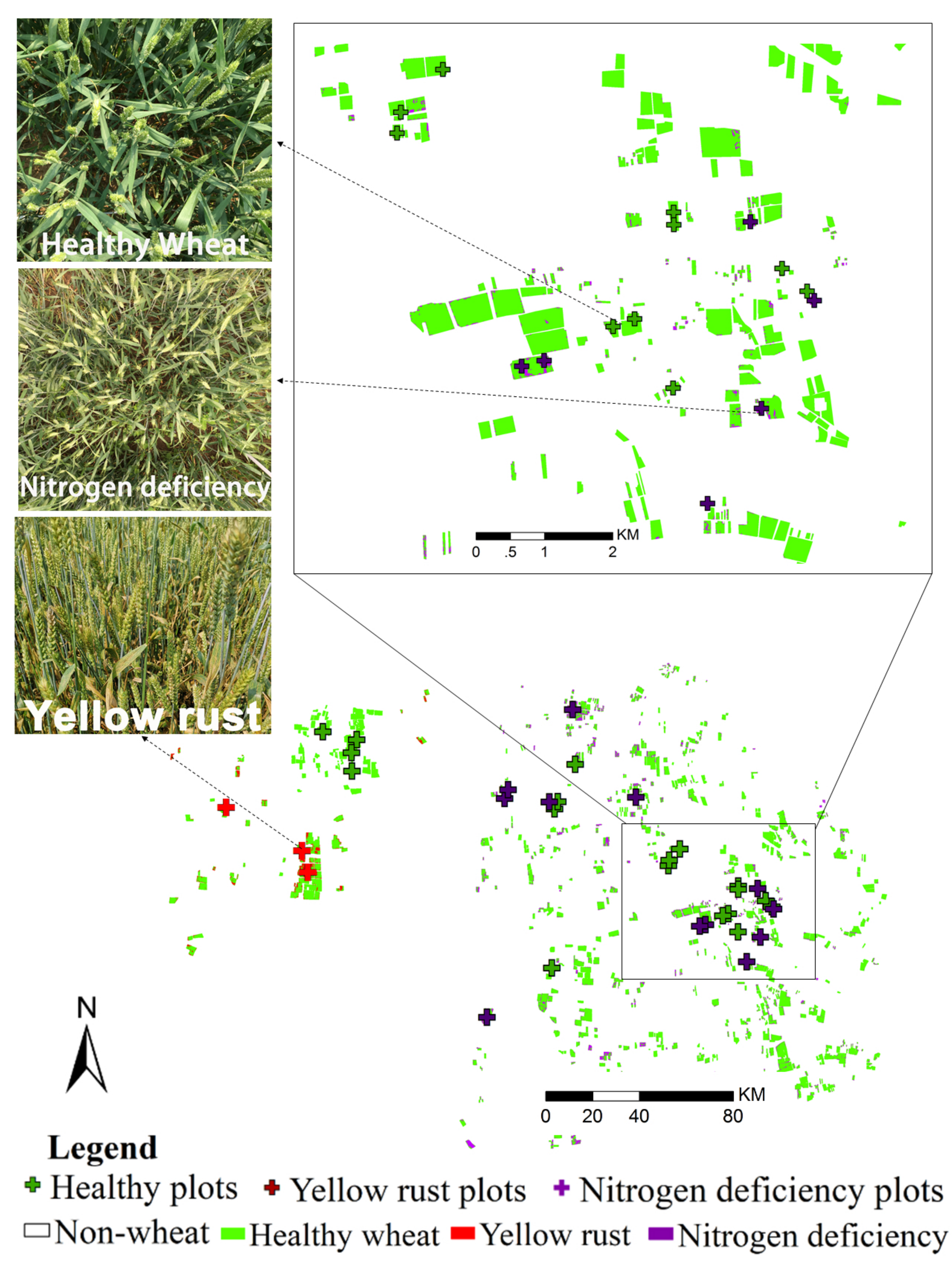}   
    \caption{ The occurrence monitoring and mapping of yellow rust in Ningqiang county, Shaanxi province (the window shows the zoom in for the classificaiton on the sub-region).}  
    \label{fig:9}  
\end{figure}

\begin{figure}[]   
    \centering  
    \includegraphics[width=5.5in]{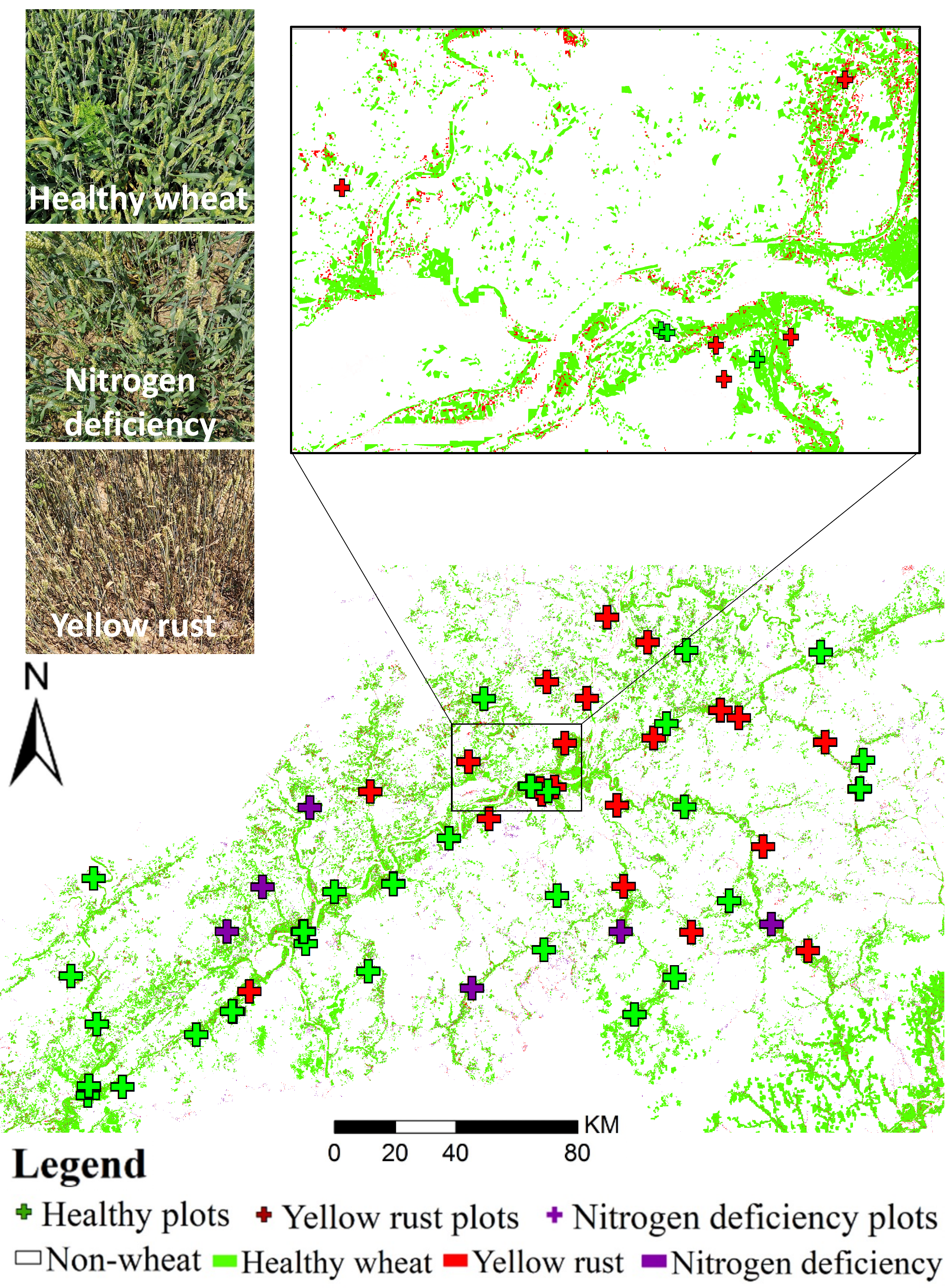}   
    \caption{ The detection and discrimination of yellow rust and nitrogen deficiency in Shunyi district, Beijing (the window shows the zoom in for the classificaitoin on the sub-region).}  
    \label{fig:10}  
\end{figure}

For the demonstration purpose, the FFCDNN-based classification map of the yellow rust and nitrogen deficiency in Ningqiang and Shunyi are respectively illustrated in Fig. \ref{fig:9} and Fig. \ref{fig:10}. The spatial distributions of yellow rust and nitrogen deficiency in Ningqiang and Shunyi are consistent with our field survey. Specifically, for  Ningqiang case, the yellow rust is mainly located around the river where ideal moisture is provided for the infestation and development of yellow rust (see the zoom-in window in Fig. \ref{fig:9}), the nitrogen deficiency is distributed around the edge of the county where the high transportation cast results in the poor fertilization management. For the Shunyi case, the nitrogen deficiency mainly occurs in the edge of the field patches (see the zoom in window in Fig. \ref{fig:10}), the yellow rust slightly occurs in the west of the study area. These monitoring results are double-checked through telephone interviews with the local plant protection department.

\subsection{ The interpretability assessment of the model}
The interpretability is one of the important matrices that measuring bias and provide explainble reason for prediction decisions from model. In this study, the interpretability assessment mainly focus on the data life in the proposed FFCDNN model and the representations of the intermediate features. 
\subsubsection{ The data life in the proposed FFCDNN model}
\begin{figure}[]   
    \centering  
    \includegraphics[width=5.5in]{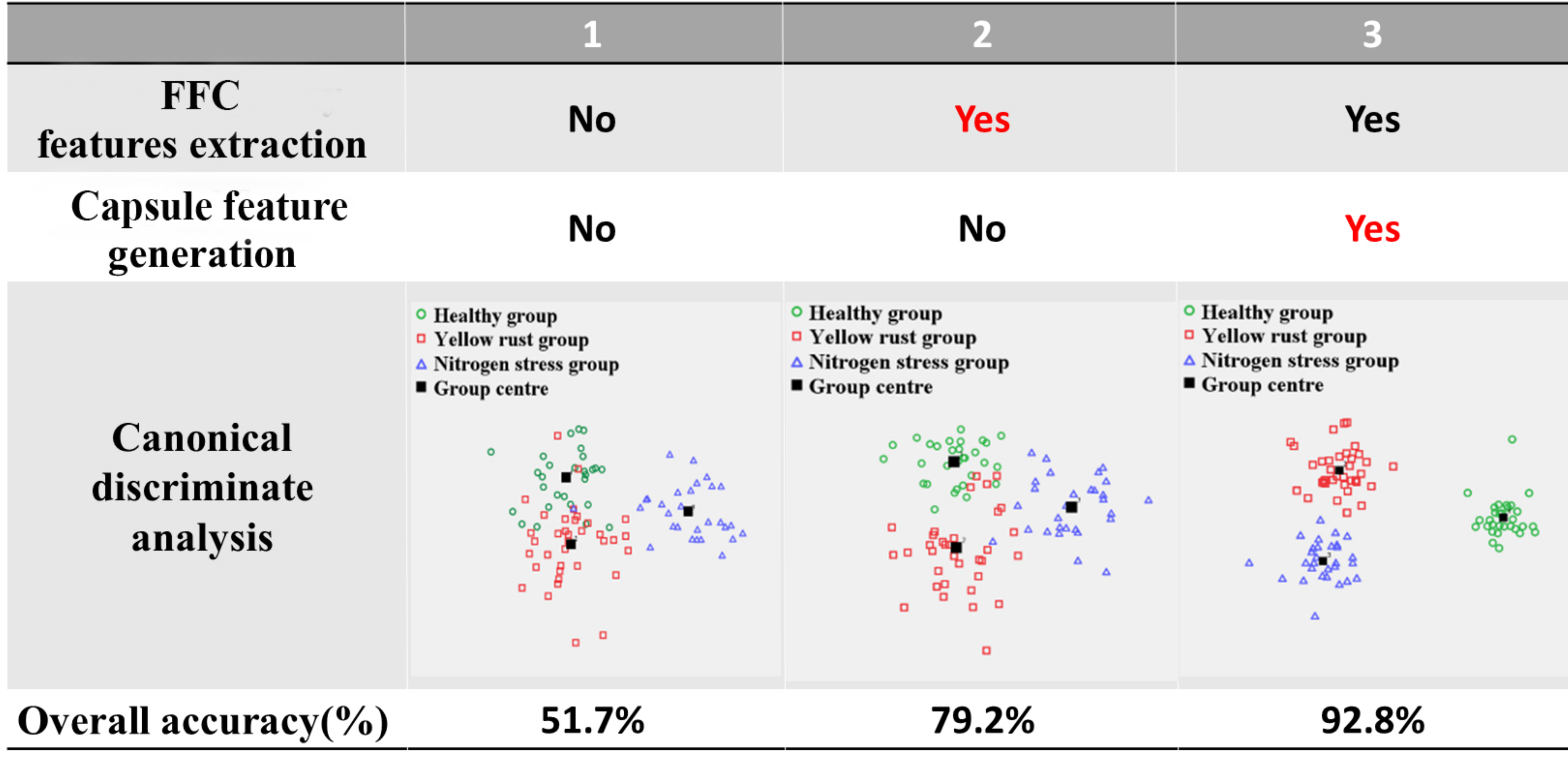}   
    \caption{ The visualization of the comparison for showing the effects of each module in FFCDNN on the canonical discriminate analysis and overall accuracy. Each column is a model with the modules in the top. The red highlight the main difference of the current model with the previous one.}  
    \label{fig:11}  
\end{figure}

In this study, two significant modules are proposed to characterize the yellow rust- and nitrogen deficiency-associated information from the Sentinel-2 time-series, thus, 1) the FFC features extraction and 2) the capsule feature generation. In order to evaluate the effects of each module on the inter-class separability, we conduct a canonical discriminate analysis to measure the clusters of the intermediate features. In the canonical discriminate analysis, the first two canonical discriminant functions are employed to establish the projective scatter plots. In addition, we gradually add the modules into the FFCDNN framework and compare their effects on classification accuracy. The visualization of the comparison is illustrated in Fig.\ref{fig:11}. \par
\textit{a.The base model without the characterized modules} \par
The base model architecture without the characterized modules is similar to a multi-layer perception (MLP), thus, the $VI_{LAI}$ and $VI_{LCC}$ time-series produced by the biological feature retrieval layer $L^{(1)}$ will directly input into the classifier $L^{(5)}$. The inter-class separability of the time-series features is shown in the second column of Fig.\ref{fig:11}, and the overall accuracy achieved by the base model is approximately $51.7\%$. \par
\textit{b. Adding the FFC layer} \par
In the FFCDNN, the FFC features extraction is the most important step to extract the yellow rust and nitrogen deficiency associated $VI_{LAI}$ and $VI_{LCC}$ frequency-domain features from the background noises. The canonical discriminate analysis indicates that, by comparison with the time-series features, the extracted frequency-domain features reveal the greater clusters between the different classes (the third column of Fig.\ref{fig:11}), and the overall accuracy reaches approximately $79.2\%$. \par
\textit{c. Adding the capsule feature encoder} \par
The capsule feature encoder is the most intelligent part of the proposed FFCDNN, which encapsulates the extracted scalar biological features into the vector features with the explicit biological representation of the target classes. The evident clusters and class edges can be figured out in the canonical projected scatter plot (the firth column of Fig.\ref{fig:11}), and the final overall accuracy reaches $92.8\%$.

\subsubsection{The representations of the intermediate features}
The primary contribution of this study is to model the part-to-whole relationship between the Sentinel-2-derived biological agents (i.e. $VI_{LAI}$ and $VI_{LCC}$) and the specific stresses, by encapsulating the scalar FFC features into the low-level class-associated vector structures. The philosophy behind the biologically composed features is that the vector features provide a hierarchical structure to represent the entanglement of the $VI_{LAI}$ and $VI_{LCC}$ fluctuations associated with yellow rust and nitrogen deficiency, and provide evidence for detection and discrimination of yellow rust and nitrogen deficiency. \par
\begin{figure}[]   
    \centering  
    \includegraphics[width=5.5in]{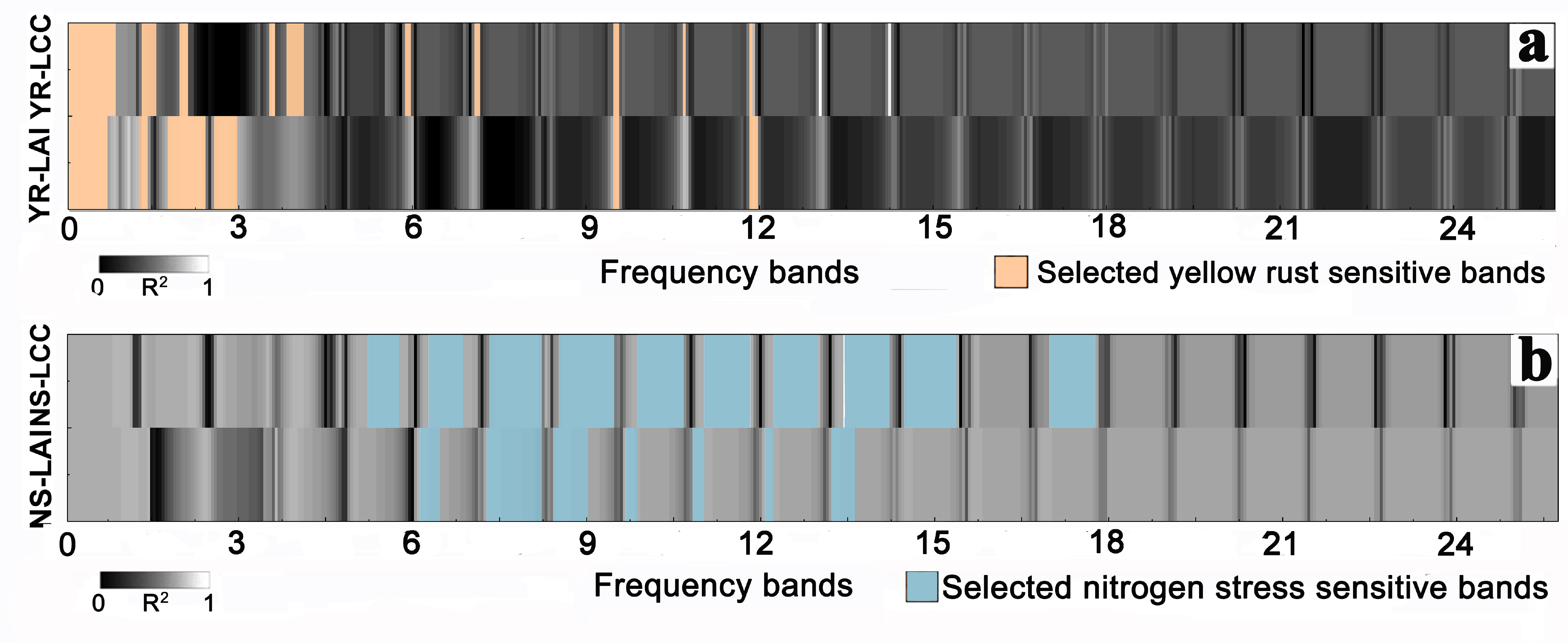}   
    \caption{ The coefficients of determination ($R^2$) between the components of the generated biological composed features and the ground measured severity of (a) yellow rust and (b) nitrogen deficiency.}  
    \label{fig:12}  
\end{figure}

The coefficients of determination ($R^2$) between the components of the generated biological composed features and the ground-measured severity of yellow rust and nitrogen deficiency are calculated based on univariate correlation analysis (see Fig.\ref{fig:12}). It’s noted that, according to Nyquist theorem, the maximum frequency component after FFT is 26 HZ, thus, the dimensionality of the generated biological composed features will be less than 52. Our results illustrate that, for yellow rust, both the $VI_{LAI}$ and $VI_{LCC}$ frequency features located in the low-frequency regions ($2 ~ 4 Hz$) highly relate with the severity levels of yellow rust, which means the host-pathogen interaction of yellow rust may induce the chronic impacts on the $VI_{LAI}$ and $VI_{LCC}$ fluctuation. These findings are in agreement with the biophysical and pathological characteristics of yellow rust that were reported in our previous study \cite{RN26}. For nitrogen deficiency, the associated $VI_{LAI}$ fluctuations are mainly located in the frequency regions of $5 ~ 15Hz$, and the associated $VI_{LCC}$ fluctuations are mainly located in the frequency regions of $6 ~ 13Hz$. Which means the nitrogen deficiency may give rise to a more acute $VI_{LAI}$ and $VI_{LCC}$ responses than that of yellow rust on the Sentinel-2 time-series. For instance, as reported in \cite{RN31}, the occurrence of nitrogen deficiency in green plants is associated with the poor photosynthesis rates, and further lead to abnormal LAI and LCC (i.e. reduced growth and chlorotic leaves). \par

\section{Conclusion}
The proposed FFCDNN model differs from existing approaches to the detection and discrimination of multiple plant stresses in the following three aspects: 1) Our model primarily considers plant biochemical information specific to the stresses. 2) The proposed Fast Fourier convolution kernel represents the first attempt to use the FFT-based kernel in a deep neural network for biological dynamic extraction from the Sentinel-2 time-series. 3) The well-designed capsule feature encoder demonstrates excellent performance in modeling the part-to-whole relationship between the extracted biological dynamics and the host-stress interaction. These three characteristics improve the interpretability of our model for decision making, akin to human experts.\par

However, two challenges persist in the practical use of the proposed implementation. Firstly, the performance of our model is inherently limited by the accurate extraction of the biochemical pre-filter. The Sentinel-2 based $VI_{LAI}$ and $VI_{LCC}$ estimations struggle to represent the real LAI and LCC values accurately, leading to the mis-estimation of the biological dynamics of specific stresses. Secondly, errors from the gap conditions and the co-registration of Sentinel-2 imagery introduce uncertainty in the modeling processes. These are primary reasons for the performance decline in the practical application of the FFCDNN. Future research will investigate whether integrating information provided by multi-source satellites into the FFCDNN framework could compensate for the LAI and LCC estimations and gap-related error, thereby further improving accuracies in detecting and discriminating yellow rust and nitrogen deficiency.\par

In conclusion, modeling the biochemical progress of specific plant stress is a key factor that influences the effectiveness of deep learning applications in the remote sensing detection and discrimination of multiple plant stresses. In this study, we proposed the FFCDNN model to analyze the stress-associated $VI_{LAI}$ and $VI_{LCC}$ biological responses from Sentinel-2 time-series to achieve multiple plant classification at the regional level. Comparisons with state-of-the-art models reveal that the proposed FFCDNN exhibits competitive performance in terms of classification accuracy, robustness, and generalization ability. \par

\section*{Conflict of Interest Statement}

The authors declare that the research was conducted in the absence of any commercial or financial relationships that could be construed as a potential conflict of interest.

\section*{Author Contributions}

YS planned the study, designed the field experiments, developed the algorithm,  and drafted the manuscript. LXH and DD reviewed, edited, conducted interviews and supervised the manuscript and lead the revision. PGM and WJH prepared and conducted interviews, reviewed and edited the manuscript and conducted interviews. ZQZ, YYL and MN provided literature reviews, HM and MD reviewed and edited the manuscript. All authors improved the manuscript by responding to the review comments. All authors contributed to the article and approved the submitted version.

\section*{Funding}
This work was supported by BBSRC (BB/R019983/1), BBSRC (BB/S020969/1), and Jiangsu Provincial Key Research and Development Program-Modern Agriculture (Grant No. BE2019337) and Jiangsu Agricultural Science and Technology Independent Innovation (Grant No. CX(20)2016).  

\section*{Acknowledgments}
The authors would like to thank Dr. Bo Liu for providing the field for our experiments in Langfang in this study.

\bibliographystyle{unsrtnat}
\bibliography{ref}  






\end{document}